\documentclass[10pt,journal,compsoc]{IEEEtran}
\ifCLASSOPTIONcompsoc
  \usepackage[nocompress]{cite}
\else
  \usepackage{cite}
\fi

\ifCLASSINFOpdf
\else
\fi

\usepackage{graphicx,amsmath,epstopdf, flushend, mathtools, amsfonts}
\begin{document}

\title{Machine Vision System for 3D Plant Phenotyping}

%Michael~Shell,~\IEEEmembership{Member,~IEEE,}
\author{Ayan~Chaudhury,
        Christopher~Ward,
        Ali~Talasaz,
        Alexander~G.~Ivanov,
        Mark~Brophy,
        Bernard~Grodzinski,
        Norman~P.A.~H\"uner,
        Rajni~V.~Patel,
        and~John~L.~Barron% <-this % stops a space
        \IEEEcompsocitemizethanks{\IEEEcompsocthanksitem Ayan Chaudhury, Mark Brophy and John L. Barron
        are with the department of Computer Science, University of Western Ontario, Canada \protect\\
E-mail: \{achaud29,mbrophy5,barron\}@csd.uwo.ca
\IEEEcompsocthanksitem Christopher Ward and Rajni V. Patel are with
              Canadian Surgical Technologies \& Advanced Robotics,
              University of Western Ontario, Canada \protect\\
E-mail: (cdwward@gmail.com; rvpatel@uwo.ca)
\IEEEcompsocthanksitem Ali Talasaz is at Stryker Mako Surgical Corporation
2555 Davie Rd, Fort Lauderdale, FL 33317, United States \protect\\
E-mail: talasaz.ali@gmail.com
\IEEEcompsocthanksitem Alexander G. Ivanov is with the Institute of Biophysics \& Biomedical Engineering,
              Bulgarian Academy of Sciences, Sofia, Bulgaria \protect\\
E-mail: aivanov@bio21.bas.bg
\IEEEcompsocthanksitem Norman P.A. H\"uner is with the department of Biology \& The Biotron Centre for Experimental Climate Change Research, University of Western Ontario, Canada \protect\\
E-mail: nhuner@uwo.ca
\IEEEcompsocthanksitem Bernard Grodzinski is with the department of Plant Agriculture,
             University of Guelph, Canada \protect\\
E-mail: bgrodzin@uoguelph.ca
}

\thanks{}}

\markboth{}%
{Shell \MakeLowercase{\textit{et al.}}: Bare Demo of IEEEtran.cls for Computer Society Journals}

\IEEEtitleabstractindextext{%
\begin{abstract}
Machine vision for plant \emph{phenotyping} is an emerging research area for producing high throughput in agriculture and crop science applications. Since 2D based approaches have their inherent limitations, 3D plant analysis is becoming state of the art for current \emph{phenotyping} technologies. We present an automated system for analyzing plant growth in indoor conditions.  A gantry robot system is used to perform scanning tasks in an automated manner throughout the lifetime of the plant. A 3D laser scanner mounted as the robot's payload captures the surface point cloud data of the plant from multiple views.  The plant is monitored from the vegetative to reproductive stages in light/dark cycles inside a controllable growth chamber. An efficient 3D reconstruction algorithm is used, by which multiple scans are aligned together to obtain a 3D mesh of the plant, followed by surface area and volume computations.  The whole system, including the programmable growth chamber, robot, scanner, data transfer and analysis is fully automated in such a way that a naive user can, in theory, start the system with a mouse click and get back the growth analysis results at the end of the lifetime of the plant with no intermediate intervention. As evidence of its functionality, we show and analyze quantitative results of the rhythmic growth patterns of the dicot \emph{Arabidopsis thaliana}(L.), and the monocot barley (\emph{Hordeum vulgare} L.) plants under their diurnal light/dark cycles.  
\end{abstract}

\begin{IEEEkeywords}
Robotic Imaging, {Arabidopsis thaliana}, Barley, 3D Plant Growth,
Multi-view Reconstruction, Diurnal Growth Pattern, Phenotyping.
\end{IEEEkeywords}}

\maketitle
\IEEEdisplaynontitleabstractindextext
\IEEEpeerreviewmaketitle

\IEEEraisesectionheading{\section{Introduction}\label{intro}}

\IEEEPARstart{A}{utonomous} and accurate real time plant \emph{phenotyping} is a quintessential part
of modern crop monitoring and agricultural technologies. Non-invasive analysis of plants
is highly desirable in plant science research because traditional techniques usually require the
destruction of the plant, thus prohibiting the analysis of the growth of the plant
over its life-cycle.
Machine vision systems allow us to monitor, analyze and produce high throughput in
an autonomous manner without any manual intervention.
Among several parameters of plant phenotyping, growth analysis is very important for
biological inference. \emph{Functional analysis} of growth curve can reveal 
many underlying functionalities of the plant \cite{leafTracking-TCBB-2015}. 
A plant's growth pattern 
can reveal different biological properties in different environmental conditions. 
For example, leaf elevation angle has impact from the amount of
sunlight \cite{diurnal-FPB-2012}. In direct sunlight, leaves
exhibit more elongation than when the plant is shaded. Similar kinds of behaviour are
exhibited for rosette size, stem height, plant surface area and volume.
Imaging techniques are very effective in terms of non-invasive and accurate analysis.
While 2D imaging techniques have been used extensively in the literature, it has some
inherent limitations. The advantage of 3D over 2D are numerous. For example, consider
the area of a leaf. If the leaf is curved, the 3D area will be significantly different from
the area computed from it's 2D image. Another restriction of 2D is that it is often difficult
to measure 3D quantities such as the surface area or the volume of a plant without doing 
error prone and potentially complex calculations, such as a stereo disparity calculation
from stereo images, to get the 3D depth information that is a precursor to these calculations.
Recently,  3D laser 
scanners are being used in many applications for studying plant phenotyping. 
As far as we know, we are the first to capture the full 3D structure of a plant
as a single closed 3D triangular mesh using a (near-infrared) laser scanner.

With the advancement of robotic technologies, automation tasks have become easier.
Apart from automating the phenotyping process, the data collection can also be
automated efficiently in real time. However, there are a number of challenges 
involved in accomplishing this, such as communication among the hardware devices, 
reliable data transfer and analysis, fault
tolerance, etc. We have developed a 3D plant phenotyping vision system
which is capable of monitoring and analyzing a plant's growth over it's entire life-cycle.
Our system has several parts. First, a gantry robot system is used for the
automation of data collection process. The robot is programmable and can be moved
around the plant by specifying a particular trajectory. A 3D laser scanner is 
the robot arm's payload. The scanner can record 3D point cloud data from a number
of viewpoints about the plant.
The robot moves from one viewpoint to another, and communicates with the scanner to
take a scan of the plant under observation. In the current setup, 
the plant is scanned six times a day from $12$ viewpoints at $30^\circ$ increments about 
$360^\circ$ (we obtain six 3D triangular meshes a day). The 12 viewpoints result in overlapping 
range data between adjacent views, allowing the
merging of all the views into a single 3D triangular mesh representing the whole plant.
Note that our laser scanner uses a near infrared beam (about $825 nm$) and 
can scan the plant in the light and in the dark equally well.

\begin{figure}[ht!]
\begin{center}
\includegraphics[width=3.0in]{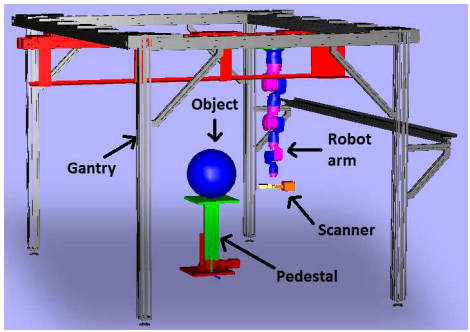}
\caption{Schematic diagram of the gantry robot system}
\label{system_schematic}
\end{center}
\end{figure}

In this paper, we report growth results and analysis for 3 weeks of the life-cycle of
wild type \emph{Arabidopsis thaliana}(L.) and barley plants.
We also attempted to measure the growth of a conifer plant, but we could not
capture the individual needle data at a sufficiently high enough resolution to
allow good range view merging (see Section \ref{limitations} for a 
complete description of this experiment).
We have grown all our plants in the indoor growth chamber
with proper environmental conditions (e.g.  light, temperature, wind, humidity, etc.).
The reason for using the Arabidopsis plant is that, this plant has used extensively 
as a model plant in biology and
was the first plant whose genome was completely sequenced \cite{DNA-nature-2000}. 
In later experiments (not yet designed or performed) we will investigate the effect
of generic modifications on the plant's growth rate as compared to the wild type.

Reconstructing a plant's 3D model from multiple views is extremely challenging
\cite{Brophy-2015, Brophy-et-al-2015, Bucksch-et-al-2013, treeGRSL-2014}.
Unlike using a rigid model like the 
Stanford bunny, reconstructing a highly non-rigid thin plant is
difficult. We use a new multi-view alignment (registration) algorithm on individual
3D point clouds obtained by our scanning to obtain a
3D triangular mesh of the growing plant. Given this mesh, we can compute
the 3D surface area and volume of the plant. We propose that area and volume make 
good plant growth metrics.

%A less-detailed preliminary report on our system \cite{ayan-crv-2015} did not
%include any biologically relevant plant growth measurements.
We have extended the basic system as reported in a conference paper \cite{ayan-crv-2015}
in several  ways. We now provide a more detailed analyses
and infer biologically relevant results for plant growth measurements. 
More specifically, we present the following new contributions in this paper.
\textbf{First}, we have made the system work throughout the lifetime of
the plants in a fully automated way, which is challenging due to the complex
nature of the whole system. \textbf{Second}, 
we propose feature point matching
of two views of a plant for rough initial registration without knowing
sensor/camera location or rotation angles. Our previous work has assumed we
had rough a priori knowledge about how much the sensor was rotated for
each view \cite{ayan-crv-2015}, now the angle between adjacent views can be arbitrary,
although they still have to overlap for the merging to work.
\textbf{Third},
we build 3D models of the plants, and 
compute surface area and volume of the reconstructed meshes
by using an efficient triangulation algorithm. \textbf{Fourth},
we analyze the growth curves of the plants 
and demonstrate the accuracy of the system to capture the well known diurnal
growth pattern of plants. 

In the next section we discuss the related literature. Then the system components are
explained in detail, followed by their integration into a complete
system and finally the operation of that system. Subsequently
the multi-view reconstruction algorithm is discussed and we show experimental
results and derive conclusions based on those results.

\section{Related Work}
In last decade there has been tremendous progress in automated plant phenotyping and plant
imaging technologies \cite{plantimaging-renaissance-COPB-2013}. Accurate phenotyping of
plants is crucial in analyzing different properties of plants in different environmental
conditions. Traditionally biologists have used na\"ive (manual) methods for plant phenotyping.
This led to low throughput and sometimes questionable accuracy.
Accuracy and throughput are major factors in mass scale analysis. To build a real
time plant phenotyping system with high accuracy, the system needs to be fully automated. 
A truly automated
system should allow a na\"ive user to operate the phenotyping process and
obtain the growth analysis as a ready-made end product. 
As many of the current phenotyping technologies
focus on software development for processing the data
\cite{htpheno-BMC-2011, phenoSoftware-2014}, automated collection
of data is also becoming state of the art \cite{scanalyzer, Subramanian-et-al-MVA-2013}
over tedious manual techniques \cite{PaulusManual-2014} .
However, most of the automated systems have their limitations. Either they are dependent on
particular types of plant(s), or on their size and geometrical structures. 
In an attempt to generalize the phenotyping process,
computer vision based system design is becoming very useful in studying plant
growth, and a body of work has been reported in that area
\cite{review-pheno-sensors-2014, PhenomicsReview-2011, FuturePhenotyping-2013}.

Among several components of phenotyping, growth analysis is extremely important
\cite{plantgrowth-importance-AoB-2008}. A plant's growth is highly affected by the environmental
conditions. Accurate measurement of a plant's growth can reveal much information which
can be useful for accelerating crop production. In recent years, different aspects of
plant phenotyping and growth measurement have appeared in the literature.

Detection and tracking of plant organs (e.g. flowers, buds, stems, leaves, fruit)
have gained the interest of many researchers.
A machine vision system for fruit harvesting  was proposed by Jimenez \emph{et al.}
\cite{Jimenez-et-al-MVA-2000}. They used an infrared laser scanner to collect data and
used computer vision algorithms to detect fruit on a plant using their colour and
morphological properties. This type of automated system is in great demand
for fruit harvesting and agricultural engineering. Automated classification of plant
organs can be useful for tracking a specific area of the plant over time. Paulus
\emph{et al.} \cite{Paulus-et-al-bmc-2013,Paulus-et-al-sensors-2014} showed a feature 
based histogram analysis method to classify different organs in wheat, grapevine and barley
plants. A segmentation algorithm to monitor grapevine growth was presented by Klodt
\emph{et al.} \cite{grapevine-BMC-2015}. A similar type of work on plant organ segmentation by 
unsupervised clustering was proposed by Wahabzada \emph{et al.}
\cite{organseg-BMC-2015}. Paproki \emph{et al.}
\cite{Paproki-et-al-bmc-2012} showed a 3D approach to measure plant growth in the
vegetative stage. Multiple images of a plant are taken from different viewpoints and the plant
mesh is generated via multi-view reconstruction. Then different organs of the plant
mesh are segmented
and parameterized. The accuracy of the parameterization is validated by
comparison with manual results. 
Golbach \emph{et al.} \cite{Golbach-et-al-MVA-2015} used a multi-camera set-up
and a plant's 3D model is reconstructed via projection matrices.
They demonstrated automatic segmentation of
leaves and stems to compute geometric properties such as area, length, etc. 
and validated the result by comparing with ground truth data 
destructively by hand. 

Other recent work focused on detecting specific patterns in a plant's leaves to 
determine the
particular condition of the plant. Analysis is performed by tracking leaves and
detecting colour properties of the leaves. However, segmentation of plant leaves
in different imaging conditions is a challenging task \cite{segmentationStudy-MVA-2016}.
Aleny\`a \emph{et al.}  \cite{tof-ICRA-2011} performed a robotic experiment for
automated plant monitoring. A camera mounted to the robot arm takes images of
the plant and the leaves are segmented. Then the robot arm
moves to the desired location to track a particular leaf. 
Dellen \emph{et al.} \cite{leafTracking-TCBB-2015} performed a $12.5$ days
experiment to analyze leaf growth of tobacco plants. They segmented the leaves 
by extracting their contours and fitted second order polynomial to model
each leaf. A graph based algorithm is used to perform tracking of leaves 
over time. Kelly \emph{et al.} \cite{Kelly-et-al-MVA-2016}
showed an active contour based model to detect lesions in \emph{Zea mays}.
This crop is widely used and detecting the lesions can be helpful to detect 
disease in the early growth stage. Xu \emph{et al.} \cite{tomato-PRL-2011} presented an
approach to detect nitrogen and potassium deficient tomatoes from the colour images
of the leaves.  Recent work on tracking leaves of rosette plants can be found in 
\cite{leaf-CEA-2015}.% *** HOW WAS THIS HELPFUL? ***
This work can be useful for measuring growth rate of a particular leaf.

Building a 3D plant model from multi-views is a challenging task. The complex
geometry of plants make the problem of 3D surface reconstruction difficult.
Pound \emph{et al.} \cite{Pound-eccv-2014, Pound-et-al-MVA-2016}
proposed a level set based approach to reconstruct the
3D model of a plant from multiple views. Their reported results are promising 
in that they show that their 3D model closely mimics the original plant structure.
Santos \emph{et al.} \cite{phenotyping-ECCV-2014} showed a feature matching based
approach to build the 3D model of a plant
using a structure from motion (SfM) algorithm. Multiple images of the plant are
taken manually and camera positions are recovered to build a 3D model of the plant.
However, the method is highly dependent on local features. 
This type of idea exploiting SfM to generate a 3D model of a plant is used in
\cite{odometry-MVA-2016} to compute plant height and leaf length accurately. 

Rhythmic patterns of a plant's growth are well studied in the biological literature
\cite{stem-diameter-1-1986,stem-diameter-2-1992,stem-diameter-3-2001}.
A system capable of detecting diurnal growth patterns can be reliably used to
monitor the growth pattern of different species in different conditions.
Imaging based techniques are becoming more popular in such analysis \cite{plantimaging-tps-2014}.
A vision based system to study the circadian rhythm of plants was presented by
Navarro \emph{et al.} \cite{circadian-sensors-2012}.
The automated system captures the diurnal growth
pattern using 2D imaging techniques. A laser scanning based 3D approach was
reported by Dornbusch \emph{et al.} \cite{diurnal-FPB-2012}, 
which shows the diurnal pattern of
leaf angle in different lighting conditions. Tracking and growth analysis
of a seedling by imaging technique was studied by Benoit \emph{et al.} \cite{cea-2014}.
Barron and Liptay \emph{et al.} \cite{Barron-et-al-1994,Liptay-et-al-1995} 
used a front and side view of a
young corn seedling imaged by a near-infrared camera to obtain growth for 1-3
days. The growth was shown to be well correlated by root temperature (Pearson 
coefficient 0.94).

Godin and Ferraro \cite{treeModel-TCBB-2010} presented a structural 
analysis of tree structures which can be useful in plant growth analysis.
Augustin \emph{et al.} \cite{Augustin-et-al-MVA-2015} modelled a mature
\emph{Arabidopsis} plant for phenotype analysis. They demonstrated extraction
of accurate geometric structures of the plant from 2D images which can
be useful for phenotyping studies of different parts of the plant.
Li \emph{et al.} \cite{4d-plant-siggraph-2013}
performed a 4D analysis to robustly segment plant parts to localize
buddings and bifurcations accurately using a backward-forward analysis
technique.

Most of the methods discussed above have several limitations. None of these
systems are designed to monitor a plant's growth for it's whole lifetime 
in an automated manner using 3D imaging technique. To the best of our knowledge, 
we are the first to report a fully automated system which operates in
near real time\footnote{Once the range scanning is complete, we start processing
the data immediately. Due to the complexity of the reconstruction algorithm,
it takes up to a $7-8$ hours to align each set of 12 views of a plant (more fully
grown plants take the full $7-8$ hours). Using a cluster of computers
provided by SharcNet, at the
end of the lifetime of the plant, the user has all the processed
results within at most four hours.
SharcNet is a supercomputing facility available to
researchers at the University of Western Ontario having
many clusters where jobs can be run in parallel, see https://www.sharcnet.ca/}
over the lifetime of a plant using laser scanning technology.
We present a novel approach to study plant growth in truly automated
manner using 3D imaging technique. The system is described in next section.

\section{System Description}
The proposed system has several parts which are integrated to make a
fully autonomous system. Each component is explained separately below.

\subsection{Gantry Robot}

A schematic diagram of the robotic system (manufactured by \emph{Schunk Inc.}, Germany)
is shown in Figure \ref{system_schematic}
comprising an adjustable pedestal and a 2-axis overhead gantry carrying a 7-DOF
robotic arm. The plant is placed on the pedestal which can be moved up and down
to accommodative different applications and plant sizes.

%The gantry robot has a $2$-axis overhead gantry through which it can move in two
%horizontal directions. A pedestal is placed at the center of the gantry, upon
%which the plant under experiment is placed. The height of the pedestal
%can be adjusted according to the size of the plant. 
%The overhead gantry supports a $7$ degrees of freedom robotic arm, which can be
%moved anywhere around the plant (shown in Figure \ref{robot_arm}).

\begin{figure}[htb!]
\includegraphics[width=3.2in, height=4.0in]{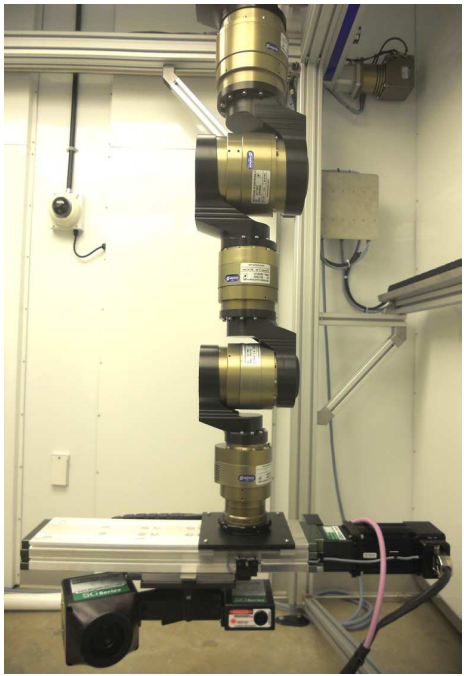}
\caption{Robot arm}
\label{robot_arm}
\end{figure}

The 7-DOF robotic arm in Figure \ref{robot_arm} provides a high level
of flexibility for controlling the position and orientation of the 3D scan head,
while the 2-axis gantry provides an extended workspace. The specification details
of the robot is shown in Table \ref{robot_table}. We can see that the accuracy
and repeatability are both 0.1 $\mu$m.

\begin{table}%[ht]
\centering
\caption{Robot Specifications}
\label{robot_table}
\begin{tabular}{| l | c | r |}
  \hline			
  Degrees of freedom &  & $7$ \\
  Dimensions & \emph{x} & $3 m$ \\
   & \emph{y} & $2 m$ \\
    & \emph{z} & $1.8 m$ \\
  Accuracy &  & $0.1\mu m$ \\
  Repeatability &  & $0.1\mu m$ \\
  Speed &  & $1$m/s \\
  Payload &  & $10$ kg \\
  \hline  
\end{tabular}
\end{table}

\begin{figure}[ht!]
\includegraphics[width=3.3in]{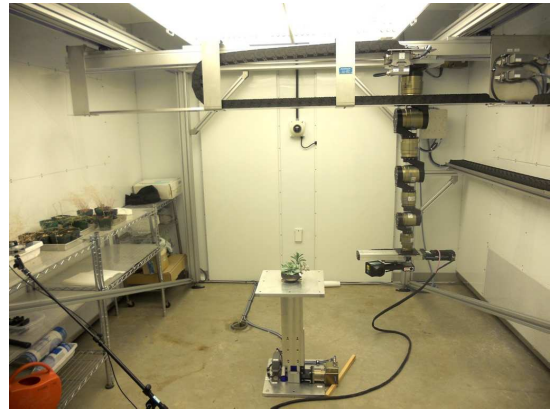}
\caption{Robot room where the experiment was performed}
\label{robot_room}
\end{figure}

We have programmed the robot to
move in a circular trajectory around the plant to take scans. Initially the
robot stays in it's home position
with the arm resting vertically downwards (Figure \ref{robot_room}).
After the initiation of the  commands, it moves from home position to the desired
location by alternating macro and micro joint movements.

% \begin{figure}[ht!]
% {Top Scanning} \\
% \includegraphics[width=3.2in, height=2.8in]{Figures/robot_arm_1} \\
% {Side Scanning}\\
% \includegraphics[width=3.2in, height=3.8in]{Figures/robot_arm_2} 
% \caption{Different scanning positions of the robot arm around the plant}
% \label{robot_arm_comb}
% \end{figure}

%Two different positions of the robot arm are shown in Figure \ref{robot_arm_comb}.
%The upper picture shows the position of the arm for scanning the plant from the top when the plant
%is in the vegetative stage. The lower picture shows the position of the arm
%used for side scanning. 
We maintained a horizontal distance of $0.56 m$ from the center of the pedestal to the
vertical axis of the robot arm, and $0.26 m$ from the pedestal plane to the scanner.
%(Figure \ref{scan_head}). 
These distances were set empirically to obtain the best scan data.

%\begin{figure}[ht!]
%\includegraphics[width=3.0in, height=1.5in]{Figures/scan_head}
%\caption{scan head distance from the plant}
%\label{scan_head}
%\end{figure}

\subsection{Scanner}

We use a \emph{SG1002 ShapeGrabber} range scanner, which is the payload of the robot arm. 
This scanner can measure dense depth maps of the visible surface of an object in
point cloud format. We also store intensity value of each point in the cloud,
although we do not currently use these values. 
The scanner uses near-infrared
light at $825$\emph{nm} (versus visible red light at $660$\emph{nm} for
most scanners) which is thought (by Grodzinski and H\"uner, two co-authors on 
this paper) to have minimal effects on plant growth 
(currently undocumented).
Different parameters
of the scanner (e.g. Field of View, laser power, etc.) are set empirically.
We have performed the whole experiment with laser power $1.0$\emph{mW}
(means the laser has a beam radius of $1.0$\emph{mm}). 
It takes about $1$ minute for a single scan to produce point cloud data
having resolution of $0.25mm$ spacing between two points. 
The scanner software Communicates over a UDP (User Datagram Protocol) link 
with the robot control software. 
Each time, after the robot stops at a scanning position, 
it communicates with the scanner to take a scan and then moves to the next position. 

\subsection{Growth Chamber}

%Growing a plant in an indoor environment can be difficult. Growing a plant like
%\emph{Arabidopsis thaliana} (L.) is challenging due to the sensitive
%nature of the plants. Although these plants can grow in unfavourable outdoor
%conditions, they need special care in the laboratory. One needs to constantly
%monitor parameters such as temperature, humidity, light, etc., otherwise it
%won't survive.
We have designed the whole robotic set-up inside a growth chamber
(manufactured by \emph{BioChambers Inc.}, Canada). The chamber is fully programmable
allowing control of the temperature, the humidity, the fan speed  and light intensity.
Also, it can be monitored remotely using a camera.
The chamber is $5.2 m^2$ % $57$ square feet 
and is equipped with a combination of $1220$\emph{mm} T5HO fluorescent lamps and halogen lamps
The whole chamber is a dedicated embedded system, and can be controlled from the
robot control software,

\section{Alignment of multi-view scans}

Multi-view alignment is a major task in building a 3D model of an object.
Pairwise registration is a crucial part in performing multi-data alignment
and this has been studied extensively in the computer vision literature
\cite{Chui-Rangarajan-2000,Tsin-Kanade-2004,Zhang-et-al-2012,Somayajula-et-al-2012}.
However, registration of thin non-rigid plant structure is very challenging and little studied.
Although Iterative Closest Point (ICP) \cite{Besl-McKay-92} and it's variants
\cite{Turk-Levoy-1994, Huber-Hebert-2003, Bouaziz-et-al-2013}
have been successful in some cases, registering highly non-rigid thin plant structures 
is still problematic.

Recently, the use of Gaussian Mixture Models (GMM) have been popular in registering two
non-rigid point sets \cite{GMM-2011}. The key idea is to represent discrete points by 
continuous probability density functions so that the problem of
minimizing discrete optimization problem can be reduced to
a continuous optimization problem.

Mathematically, a Gaussian mixture model can be 
stated as the weighted sum of M component Gaussian densities:
\begin{equation}
p(\mathbf{x}|\lambda) = \sum_{i=1}^{M} w_i\  g(\mathbf{x}|\mathbf{\mu}_i,\mathbf{\Sigma}_i),
\end{equation}
where $\mathbf{x} $ is a $d$-dimensional vector, 
$w_i (i=1,...,M$) are weights of each mixture $g$ having mean
$\mathbf{\mu}_i$ and covariance $\Sigma_i$, and $\lambda = \{w_i, \mathbf{\mu}_i, \mathbf{\Sigma}_i\}$
are the parameters.
The registration problem can be formulated as the  minimization of the discrepancy 
between two Gaussian mixtures by minimizing the following cost function \cite{GMM-2011}:
\begin{equation}
d_{L_{2}}(\mathcal{S},\mathcal{M},\Theta) = \int  (gmm(\mathcal{S}) - gmm(\mathcal{T}(\mathcal{M},\Theta)))^2  \: \mathrm{d}x
\end{equation}
where $\mathcal{M}$ is the model point set and $\mathcal{S}$ is the scene (data) point set. 
The function $gmm(\mathcal{P})$ denotes the Gaussian
mixture density constructed from $\mathcal{P}$. The aim is to find
parameters $\Theta$ which minimizes
the above cost function using \textit{$L_2$ norm} as the distance measure.
Myronenko \emph{et al.} presented a Coherent
Point Drift (CPD) algorithm \cite{CPD-2010}
where the center of Gaussians are moved together for registering
two point sets. The method is state of the art and has been applied to many
applications. These methods work reasonably well for pairwise registration but
aligning multi-data using them is problematic. 

We use an extension of CPD for aligning multi-datasets, proposed by
Brophy \emph{et al.} \cite{Brophy-et-al-2015}. 
Given two point clouds, $\mathbf{X}=(\mathbf{x}_1,\mathbf{x}_2,...,\mathbf{x}_m)^T$
and $\mathbf{Y}=(\mathbf{y}_1,\mathbf{y}_2,...,\mathbf{y}_n)^T$,
in general for a point $\mathbf{x}$, the GMM probability 
density function will be:
\begin{equation}
p(\mathbf{x}) = \sum_{i=1}^{M+1} P(i)p(\mathbf{x}|i),
\end{equation}
where
\begin{equation}
p(\mathbf{x}|i) = \frac{1}{(2\pi\sigma^2)^{d/2}}\  exp[-\frac{||\mathbf{x}-\mathbf{y}_i||^2}{2\sigma^2}].
\label{eqn_CPD_2}
\end{equation}
Instead of maximizing the GMM posterior probability, the negative log-likelihood function 
can be minimized to obtain the optimal alignment:
\begin{equation}
E(\theta,\sigma^2) = - \sum_{j=1}^{N} log \sum_{i=1}^{M+1} P(i)p(\mathbf{x}_j|i),
\label{eqn_CPD_3}
\end{equation}
where
\begin{equation}
P(i|\mathbf{x}_j) = P(i) p(\mathbf{x}_j|i) / p(\mathbf{x}_j).
\label{eqn_CPD_4}
\end{equation}
Then the Expectation Maximization algorithm is used iteratively to optimize the cost function. 

Inspired by the work in rigid registration by Toldo \emph{et al.} \cite{Toldo2010},
an ``average'' scan is constructed, to which we register all
other scans.  For a scan $X$, 
we find the set of points that are the Mutual Nearest Neighbours (MNN) to a point in the scan, and
then we calculate a scan that is composed of the calculated centroids from each point.
Once the initial registration is complete, we use CPD in conjunction with MNN 
to recover the non-rigid deformation field that the plant undergoes between the 
capture of each scan. At this point, the scans
should be approximately aligned to one another. We then construct 
the centroid/average scan and then register to it.

Basically, the method is a two step process, beginning with aligning the scans approximately. We 
then register a single scan to the ``average'' shape, constructed from all other scans, and  
update the set to include the newly registered result. We perform the same process with 
all other sets of scans.
In this way, we avoid accumulation of merging error.
In general, CPD alone is effective in registering pairs with a fair amount of overlap, 
but when registering multiple scans, our 
method achieves a much better fit than CPD by itself, utilizing sequential pairwise
registration.

\subsection{Rough Initial Alignment}

Although the multiple view registration algorithm works well in aligning
different scan data, registering two views with huge rotation angle difference
can pose difficulties. Unfortunately, 
after decades of research on point cloud registration,
state of the art algorithms fail when the views are not roughly aligned.
The problem is more challenging for the cases of complex plant structures
due to occlusion and local deformation between two views
\cite{Bucksch-et-al-2013, treeGRSL-2014, ayan-GRSL}. Note that the rough initial 
alignment alone is not sufficient to register two views, and this is a
pre-processing step of the actual registration algorithm as discussed in the 
previous section.

One approach to estimate the 
rough alignment of two views is to find corresponding feature points.
However, because of the complex structure of the plants, finding
repeatable features is extremely hard \cite{leafTracking-TCBB-2015} 
and typical feature point matching algorithms fail.
Junctions are strong features for plant-like structures. 
We adopt the idea of using junction point of branches as feature points
and then match these features \cite{ayan-GRSL}.
However, the idea in \cite{ayan-GRSL}
for detecting junction points does not consider occlusion in leaves. This results
in false feature point detection, which may not be repeatable in two views. We
extend our algorithm \cite{ayan-GRSL} using a simple but effective
density clustering technique as discussed below. 
Lin et al. have proposed a similar idea \cite{code-PAMI-2017}.

\subsubsection{Feature Clustering}

The basic idea of our junction detection algorithm \cite{ayan-GRSL} is to first extract
local neighbourhood around every point using \emph{kd}-trees and perform a 
statistical dip test to determine the non-linearity in the data.
Then the branches are approximated by fitting straight lines to the point cloud. 
Finally straight line equations are solved
to determine if they intersect in the local neighborhood.
This approach results
in detection of multiple feature points around the junction, because all the
points around a small neighbourhood at the junction are potential candidates
of true junction points, from which the best candidate is picked up by
non-maximal suppression of the dip value. But this idea also detects
false feature points as junctions in occluded leafy areas.
We handle this problem by applying density based clustering to extract
the true junctions and filter out the rest of the feature points \cite{DBSCAN-1996}. 
 
\begin{figure*}[ht!]
\includegraphics[width=6.0in, height=6.0in]{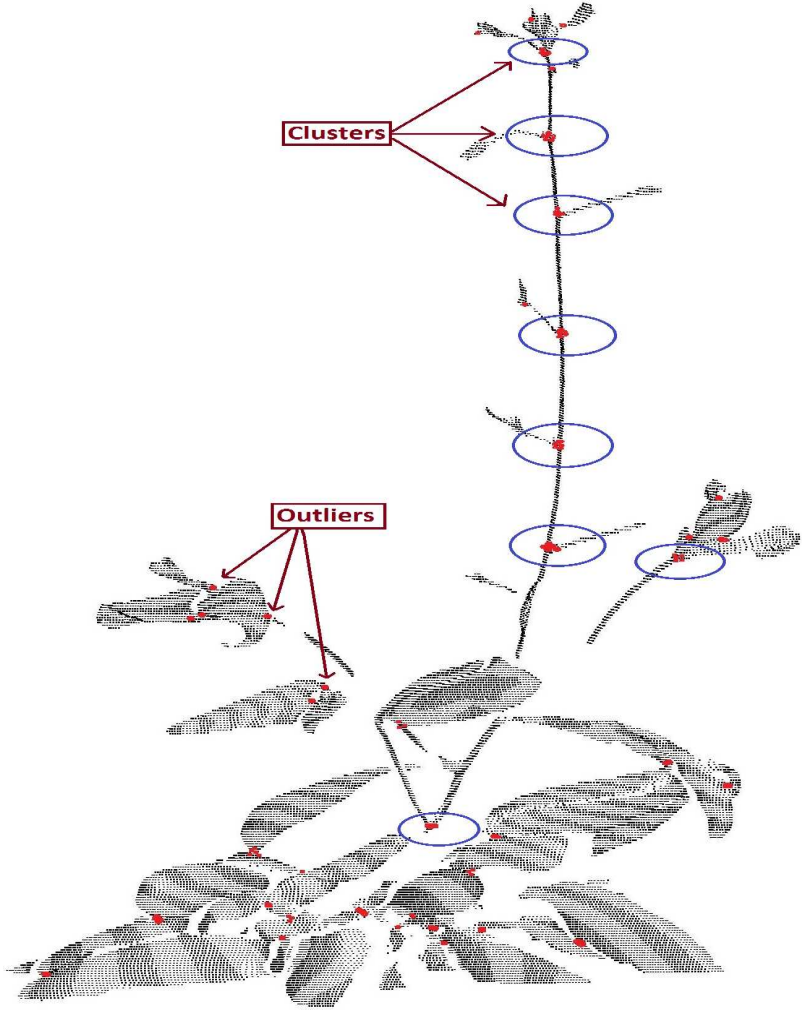}
\caption{Demonstration of feature clustering. Red dots represent junction
feature points \cite{ayan-GRSL}. Density clustering algorithm \cite{DBSCAN-1996}
detects clusters at true junctions (denoted by blue circle) and treats
false feature points as outliers.}
\label{dbscan}
\end{figure*}

The idea of density based clustering is to find groups of points 
that are denser than the remaining points. As true junction features tend to
appear with higher density than false junctions, we cluster the
detected feature points to find the cluster of points that are formed
at true junction points. We use the density based clustering
algorithm proposed by Ester \emph{et al.} \cite{DBSCAN-1996}. The algorithm does
not need the number of clusters to be known in advance (unlike \emph{k}-means) and
can perform clustering in the presence of large number of outliers. Finally, 
we compute the centroid of each cluster to find the true junctions and
match these features using our subgraph matching technique 
\cite{ayan-GRSL}. Figure \ref{dbscan}
illustrates this idea. It shows a single view of a plant having occlusions.
Red dots represents feature points detected by our junction detection algorithm
\cite{ayan-GRSL}. The clustering algorithm detects clusters around true junction 
points (denoted by blue circles) and discards the remainder of the points as outliers.

\section{System Integration}

\begin{figure*}[ht!]
\begin{center}
\includegraphics[width=5in, height=3in]{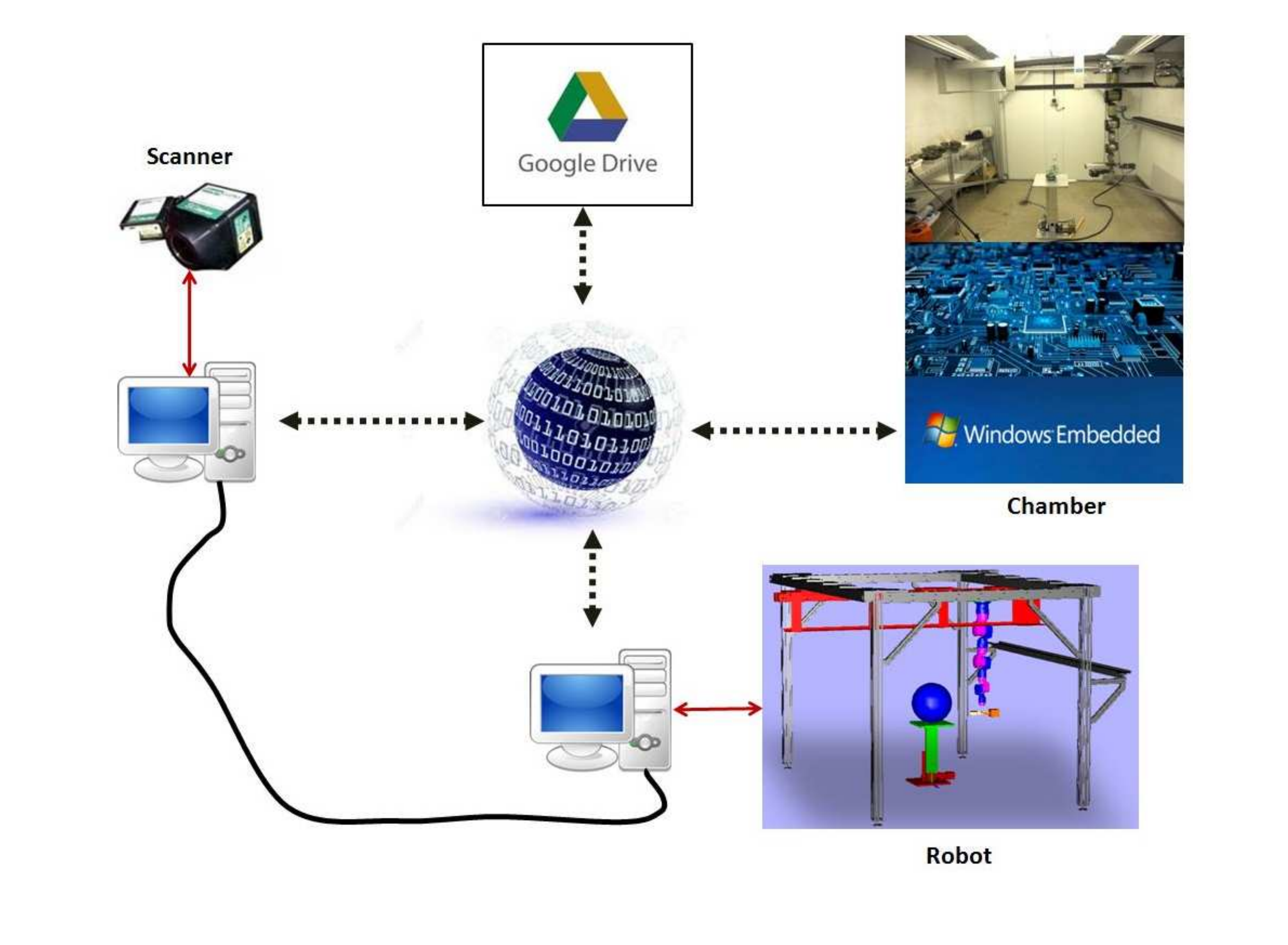}
\caption{High level view of the system}
\label{system}
\end{center}
\end{figure*}

We integrate the chamber, robot and scanner. The system components
and their connections are shown in Figure \ref{system}. The robot and scanner are 
operated from different computers which communicate over a dedicated UDP link.
The chamber  is accessed remotely over the internet. 

Communication between the chamber and robot operation needs to be done frequently.
While scanning a plant, we need to shut down or significantly decrease the
fan speed inside the chamber (at present we just shut it down completely),
otherwise the scan data will be erroneous 
as the plant will be jittering (this makes
the multiple view reconstruction problem extremely difficult). Also, 
experimenting in different lighting conditions (short day versus long day) needs
communication between the chamber and the scanning schedule of the robot. 
Before starting the experiment, chamber parameters are set according to the need of the
application. During the experiment before a set of scans, the robot communicates with the
chamber, turns off the fan (and light if needed), and restores
the default chamber settings when the scan is completed. This process is repeated
at each scan.

\begin{figure}[ht!]
\includegraphics[width=3.0in, height=2.5in]{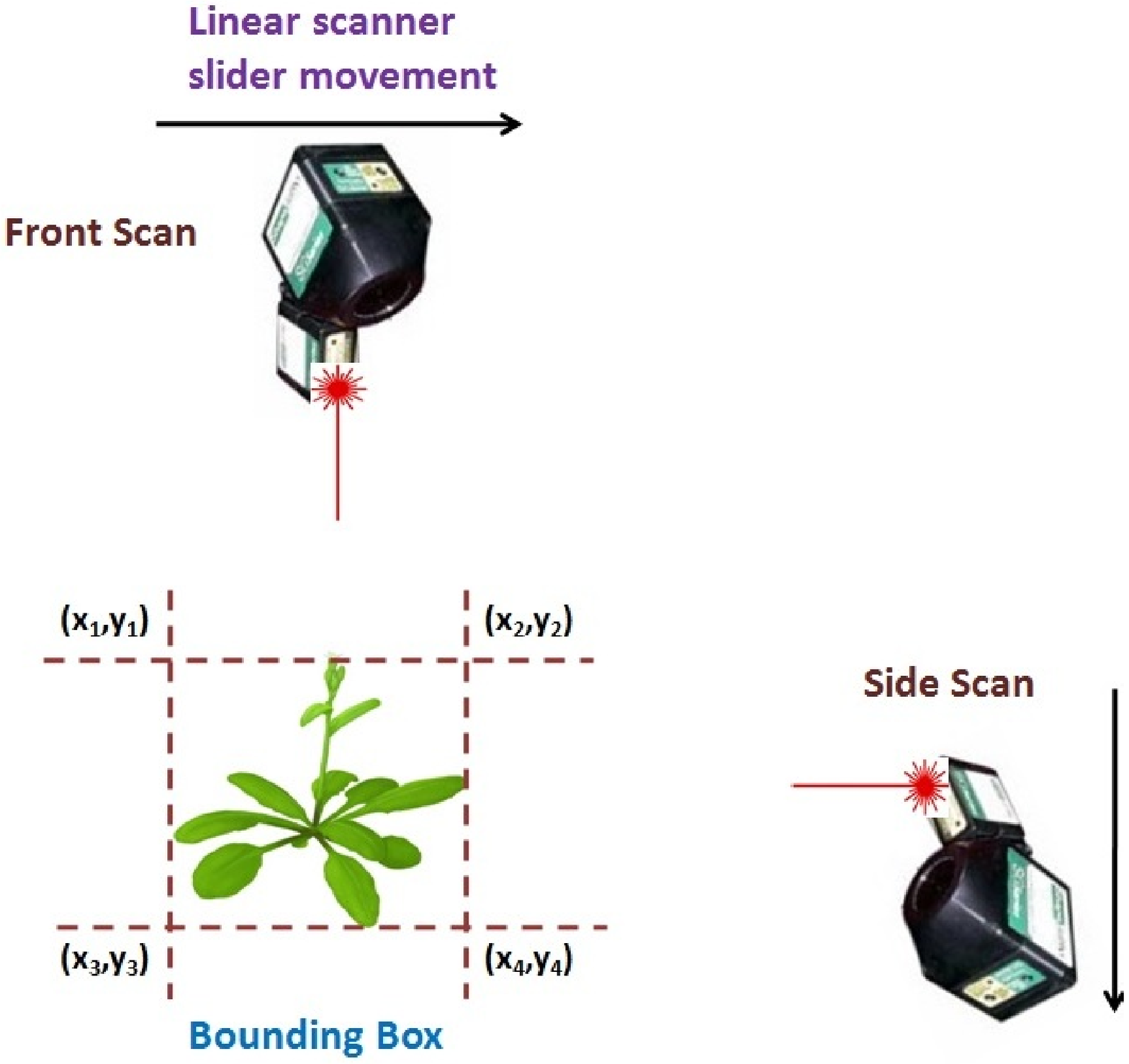}
\caption{Computing the bounding box to determine the center of the plant}
\label{bb}
\end{figure}

As the size or dimension of the plant is not known in advance, determining the scanning
boundaries to enclose the whole plant needs to be done dynamically during 
each scan. Moreover, as the plant grows, it may lean towards a particular direction,
which requires the scanner position to be adjusted accordingly. We perform a simple
bounding box calculation before performing each scan (Figure \ref{bb}). Before doing
the actual scanning, a pre-scan procedure is performed from 2 directions (front and
side as shown in the figure). 
From these scans, the centre of the bounding box of the plant is approximated
and used to update the centre of rotation for the circular scanning trajectory.
Once the plant centre has been determined, the system causes the gantry and robotic
arm to translate and rotate the scan head to specified discrete positions around
the plant. 

\begin{figure}[ht!]
\includegraphics[width=2.5in, height=2.3in]{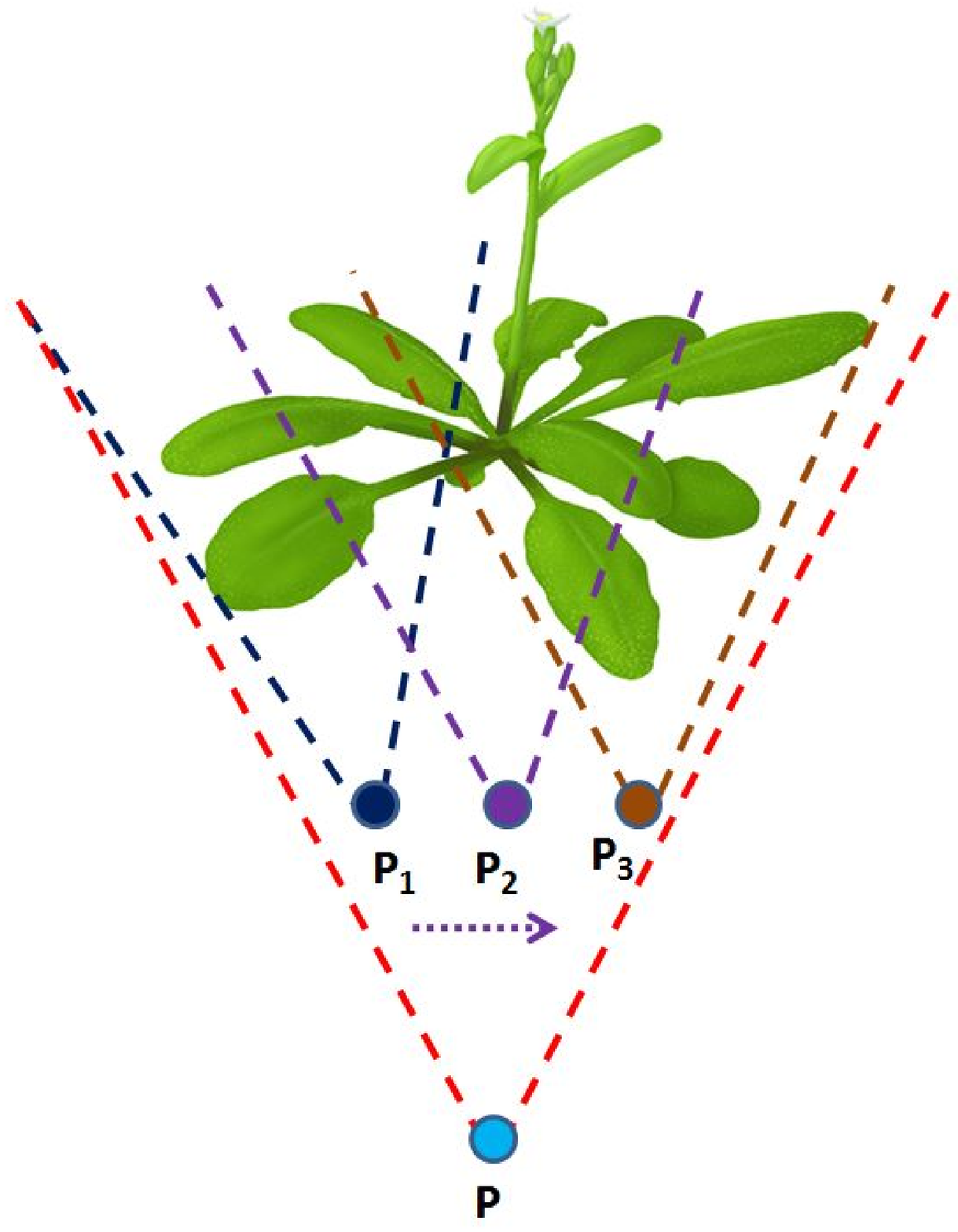}
\caption{Scanner Field of View (FOV) to enclose the whole plant. When the plant
is big, one scan can't capture the whole plant and multiple scans are required.
This is done automatically.}
\label{fov}
\end{figure}

At each scanning position, the system waits 10 seconds to allow the plant and the
scanner to settle before initiating a scan. Once a scan has finished, the resulting
scan data are analyzed to ensure that the plant has been fully captured (i.e. there
is no clipping via a bounding box calculation).
Sometimes the scanner FOV is not wide enough to capture the full width of the plant
due to a limited travel of the scanner's linear stage ($0.2m$).
In that case, an extra scan of that view is automatically made
by sliding the scanner in a sideways direction. 
From empirical observations, for plants like Arabidopsis, no more than
3 partial scans are needed to enclose a single view as illustrated in Figure \ref{fov}.
Usually, if the plant is not too wide, a single view from $P$ is sufficient. Otherwise,
we perform side scans at positions $P_1$, $P_2$ and $P_3$.
We want to emphasize that all scanning, including this side scanning is fully
automated in our system. Currently, we have not allowed for upwards scanning, where
the plant could grow out of range by growing too high (this has never happened in all our
scanning experience).

We operate the whole system from a single GUI. 
%The user controls everything
%from the GUI (Figure \ref{systemGUI_1}). 
Once the system is started, theoretically,
it can continue scanning and processing until the experiment is terminated.
We say ``theoretically'' here as some unexpected events, such as network
failure, can occur. Network failure requires human intervention to fix.

%\begin{figure}[ht!]
%\begin{center}
%\includegraphics[width=3.2in]{Figures/TestGUI_3}
%\caption{System GUI or user interface. The user can select the parameters according
%to the need of the experiment, clicks to start the system and gets all the results at
%the end of the lifetime of the plant.}
%\label{systemGUI_1}
%\end{center}
%\end{figure}

\section{Experimental Results}

First we have performed a 21 day experiment with a wild type Arabidopsis plant
(after which the plant was falling down) with a $12/12$ hours light/dark cycle at the
temperature of $25^\circ$C and a light intensity of $250\mu mol$ photons  $m^{-2}s^{-1}$.
We also performed another 15 day experiment with barley under the same conditions.
Although the algorithms used in the different stages are not new, the
main challenge was to make the whole system work continuously for the
life-cycle of the plant. 
We encountered several problems while integrating the system parts (robot,
chamber, scanner, etc). As the robot and the chamber needed to communicate frequently
(every $4$ hours as we are performing $6$ scans per day, throughout the life of
the plant), there were issues of lost communication due to network failure
and infrequent hardware faults (e.g. problems
in electronic chips at robot joints, etc).

\subsection{Merging of multi-view plant point cloud}

\begin{figure}[ht!]
\begin{center}
\includegraphics[width=2.5in,height=5.0in]{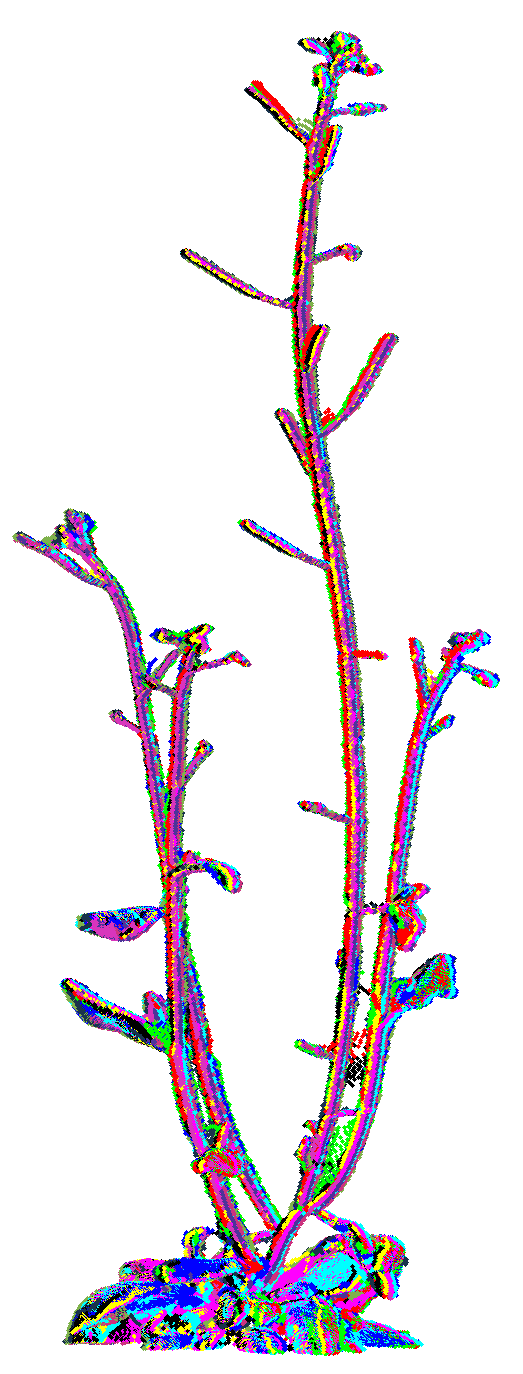}
\caption{Reconstructed Arabidopsis plant point cloud (different colors indicate
different scans)}
\label{plant_rec}
\end{center}
\end{figure}

\begin{figure}[ht!]
\begin{center}
\includegraphics[width=2.5in,height=5.0in]{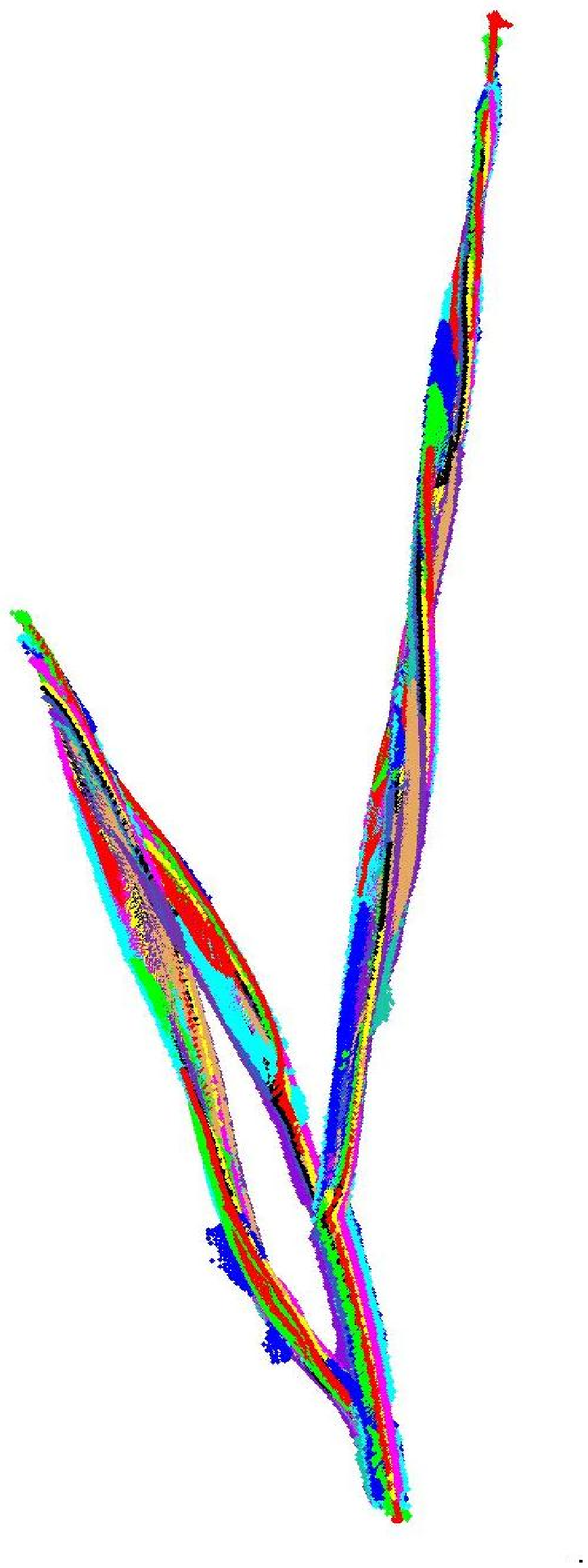}
\caption{Reconstructed Barley plant point cloud (different colors indicate
different scans)}
\label{plant_rec_barley}
\end{center}
\end{figure}

Using the reconstruction algorithm discussed above, we performed alignment of the multi-view
point cloud data of the growing plant over time. 
We used Sharcnet
computing machines for simultaneous processing of large amount of data while the scan
data was collected
throughout the life of the plant. The merged point cloud data for fully grown Arabidopsis
plant (day $20$\textsuperscript{th})
and the barley plant
(day $15$\textsuperscript{th}) are shown in Figures \ref{plant_rec} and \ref{plant_rec_barley}
respectively. Each colour represents different scans. 

\subsection{Polygonal mesh formation}
Once we have the aligned point cloud from the multi-view data, it needs to be triangulated
in order to compute the mesh surface area and volume. 
Accurate triangulation of point cloud data is a challenging problem.
An efficient triangulation should represent all the details of
the shape of the object. Triangulation of plant structures is more challenging
due to the thin branches. Although Delaunay triangulation is typically used
for modeling a surface, the algorithm does not produce good result for
plant structures. We used the $\alpha$-shape algorithm \cite{alpha-shape-1983} for
triangulation. The algorithm works well when its parameters are properly tuned
\footnote{We use the MatLab code for the $\alpha$-shape algorithm on
the MatLab file exchange website, 
https://www.mathworks.com/matlabcentral/fileexchange/}.

\subsection{$\alpha$-Shape Triangulation}
Let $P = \{p_1,\cdots,p_n\} \subset \mathbb{R}^d$ be a set of points, which
are called \emph{sites}. A \emph{Voronoi diagram}
is a decomposition of $\mathbb{R}^d$ into convex polyhedra. Each region
or \emph{Voronoi cell} $\mathcal{V}(p_i)$ for $p_i$ is defined to be the set of
points $x$ that are closer to $p_i$ than to any other site. Mathematically,

\begin{equation*}
\mathcal{V}(p_i) = \{x \in \mathbb{R}^d~~|~~||p_i - x || \leq ||p_j - x ||~ \forall j \neq i \},
\end{equation*}

\noindent where $|| . ||$ denotes the Euclidean distance. 
The \emph{Delaunay triangulation} of P is defined as the dual of
the Voronoi diagram. 

\begin{figure*}[ht!]
\includegraphics[width=7.0in,height=6.0in]{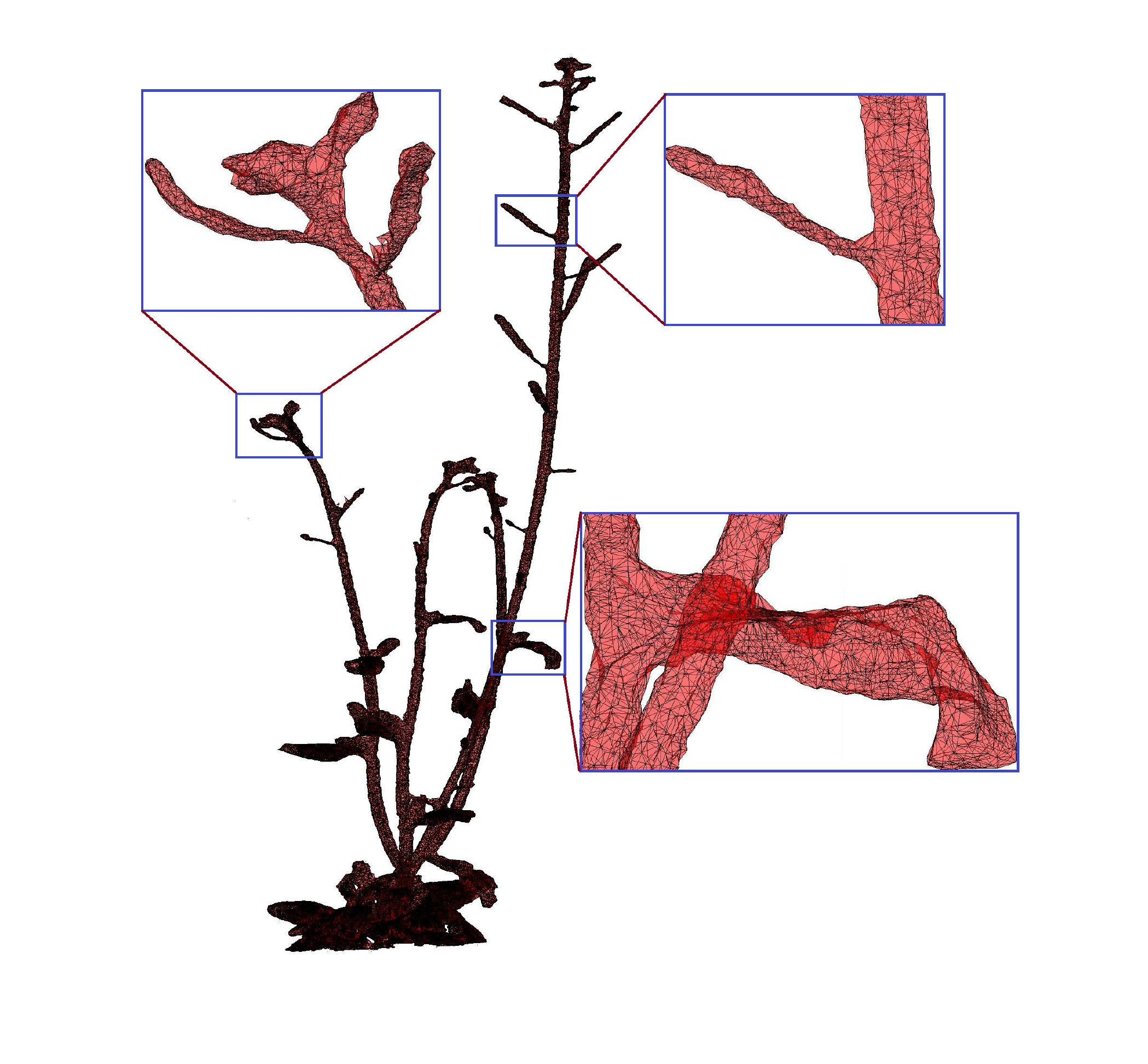}
\caption{Triangulated Arabidopsis plant data}
\label{alpha_original_arabidopsis}
\end{figure*}

\begin{figure*}[ht!]
\includegraphics[width=7.0in,height=6.0in]{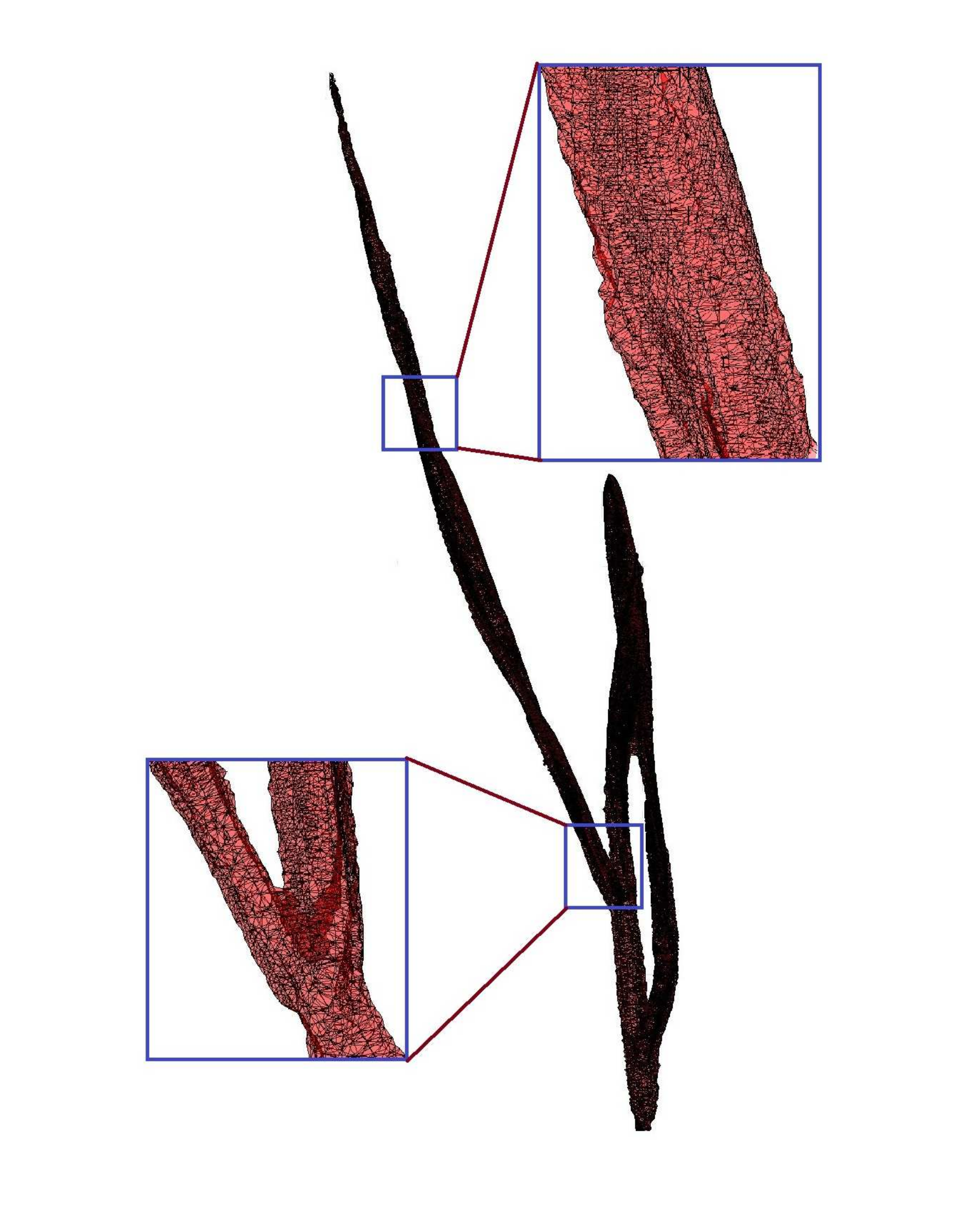}
\caption{Triangulated Barley plant data}
\label{alpha_original_barley}
\end{figure*}

The \emph{$\alpha$ complex} of $P$ is defined as the Delaunay
triangulation of $P$ having an empty circumscribing sphere with squared radius
equal to or smaller than $\alpha$. 
The $\alpha$ shape is the domain covered by alpha complex.
If $\alpha = 0$, the $\alpha$-shape is the point set $P$, and for $0 \leq \alpha \leq \infty$,
the boundary $\partial \mathcal{P}_\alpha$ of the $\alpha$-shape is a subset
of the Delaunay triangulation of $P$.
The main idea of the algorithm is that the space generated by any point pairs can be
touched by an empty disc of radius $\alpha$. The value of $\alpha$ controls the level
of detail in triangulation.
The algorithm is simple and effective. We have empirically chosen
$\alpha = 0.6$ for the plants. Also note that we have already performed ``smoothing"
of the point cloud while applying Gaussians, so we do not need further surface
smoothing (such as Poisson smoothing). Example results are shown in Figures 
\ref{alpha_original_arabidopsis} and \ref{alpha_original_barley}. 
The rectangular cutouts
show some smaller portions of the plants at higher resolution.
The surface area is simply the summation of area of each triangle in the mesh and
volume can be computed using the technique described by Zhang \emph{et al.}
\cite{meshVol-ICIP-2001}.
The ``loop'' structure is due to the angle by which
the barley plant is viewed, some of parts of the plant occludes other parts.


\subsection{Biological Relevance}

\begin{figure*}
\includegraphics[width=7in, height=4.0in]{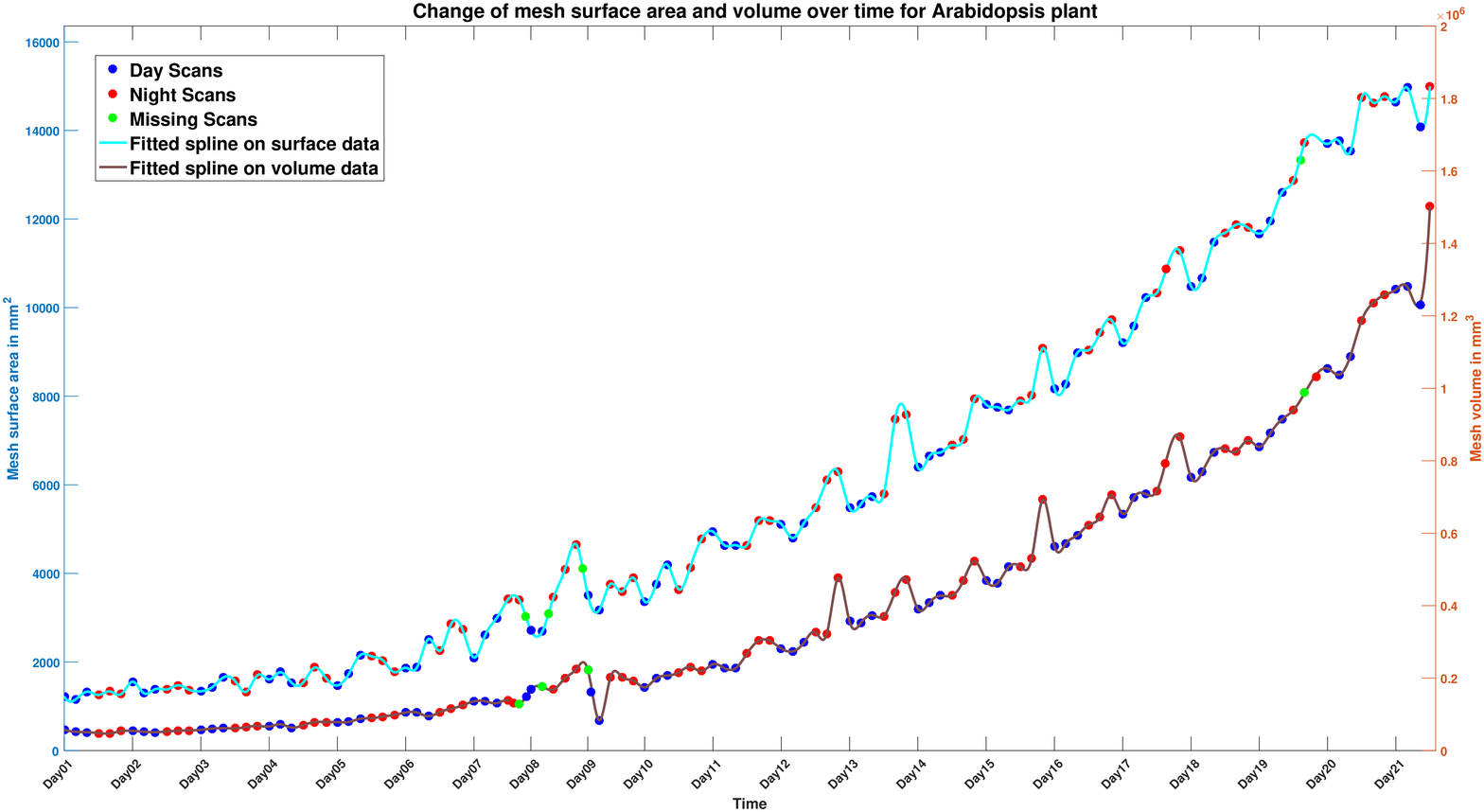}
\caption{Diurnal growth pattern of mesh surface area and volume for the Arabidopsis plant.
The red dots represent night time scans, the blue dots represent day time scans and the four
green dots represent missing scan data. A spline is fitted to both surface and
volume scan data (shown in different colours). The $y$-axis in the left and right hand side
represents the range of surface area and volume data respectively.}
\label{arabidopsis_combinedGraph}
\end{figure*}

\begin{figure*}
\includegraphics[width=7in, height=4.0in]{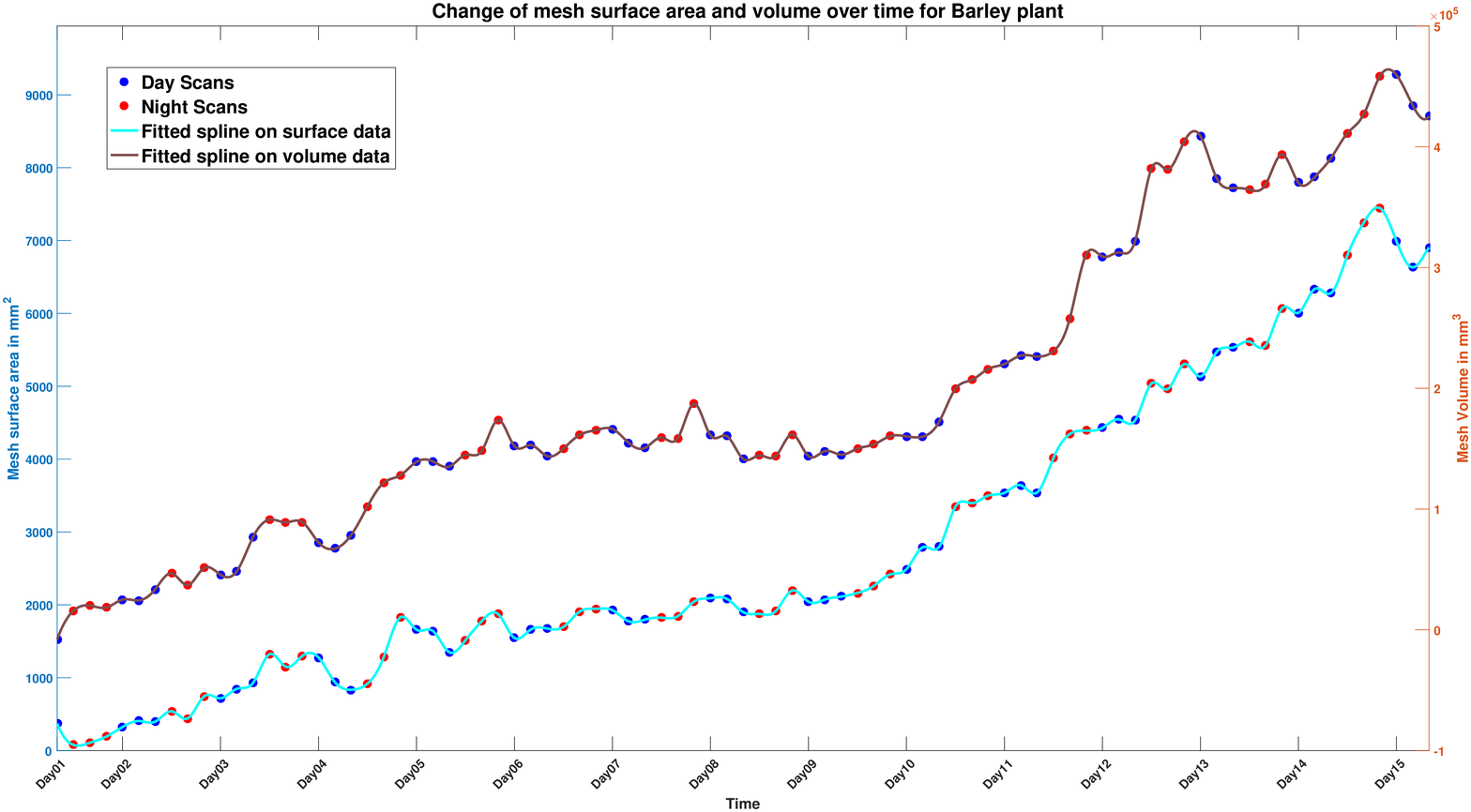}
\caption{Diurnal growth pattern of mesh surface area and volume for the barley plant.
The red dots represent night time scans while the blue dots represent day time scan data. 
A spline is fitted to both surface and
volume scan data (shown in different colours). The $y$-axis in the left and right hand side
represents the range of surface area and volume data respectively.}
\label{barley_combinedGraph}
\end{figure*}

The fact that plants grow mostly at night and shrink in the day time is well known
\cite{diurnal-Nature-2011}.
It is observed that the changes in stem diameter depends on the lighting
conditions \cite{stem-diameter-1-1986, stem-diameter-2-1992, stem-diameter-3-2001}.
While the diurnal nature of plant growth can involve changes in stem length, 
width, diameter, leaf surface area, we have observed the diurnal pattern in both 
volume and surface area of the plant. The mesh surface area and volume are plotted 
against time
in the same graph for Arabidopsis plant in Figure  \ref{arabidopsis_combinedGraph}.
A similar plot for barley plant is shown in Figure \ref{barley_combinedGraph}.
In the graphs, red dots represent night time scans and blue dots represent day time scans.
As we have $6$ scans per day, there are $3$ blue dots followed by $3$ red dots in the 
graph. For the Arabidopsis experiment we had $4$ scans missing due to networking problems.
These missing data are generated by taking the average of previous and next scan data.
These are shown as green dots in Figures \ref{arabidopsis_combinedGraph} and
\ref{barley_combinedGraph}.

It can be noticed from the {\bf growth curves} that the plants exhibited more growth in the night 
time than in the day time, which supports the biological relevance of diurnal growth pattern
of plants. Finally, note that the changes of volume are greater than the changes of surface area
in the later period of the growth cycle (this is logical as volume grows faster that area).

\begin{figure*}[ht!]
\begin{center}
\begin{tabular}{c c}
\includegraphics[width=3.45in, height=2.25in]{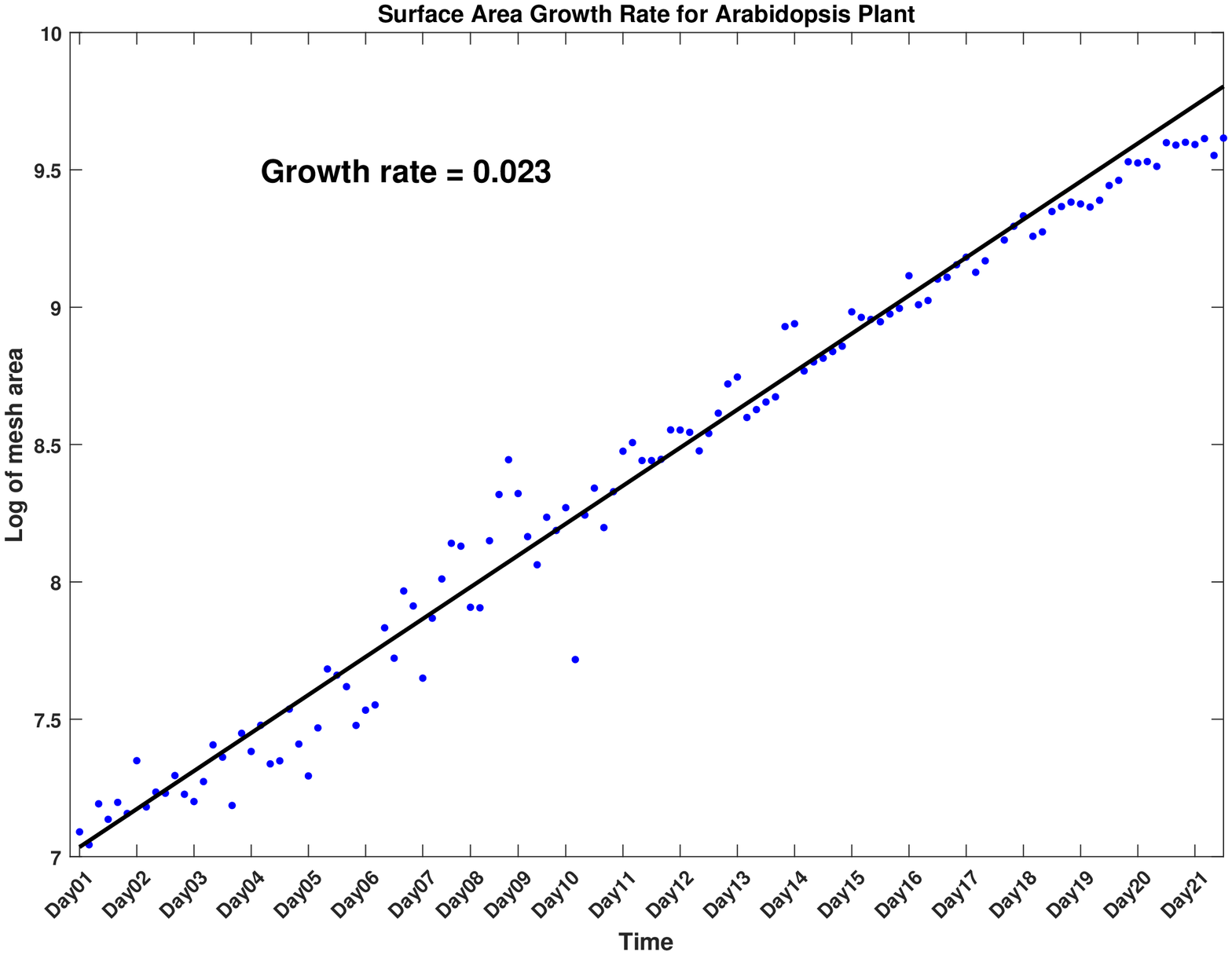} &
\includegraphics[width=3.45in, height=2.25in]{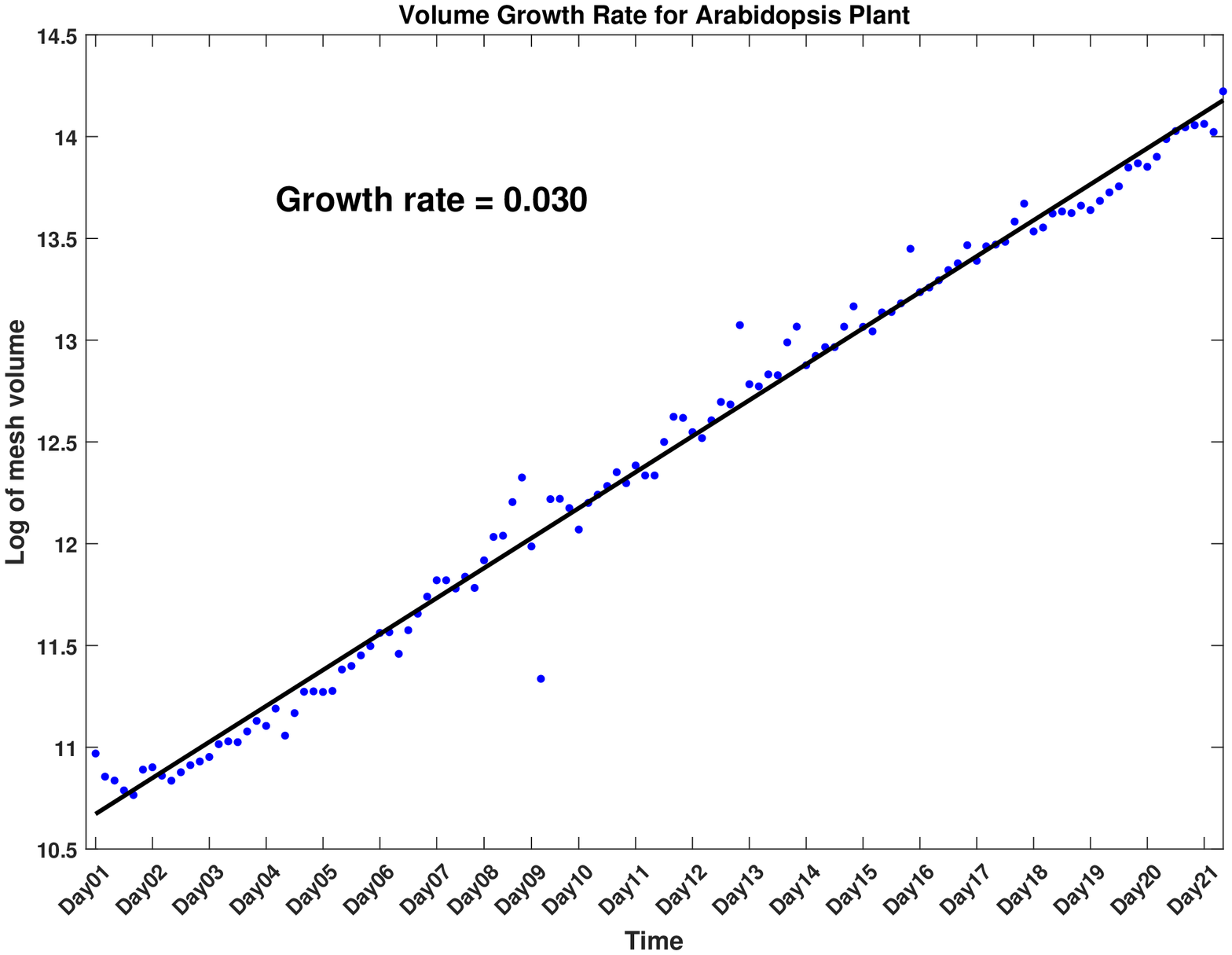} \\
(a) & (b) \\
\includegraphics[width=3.45in, height=2.25in]{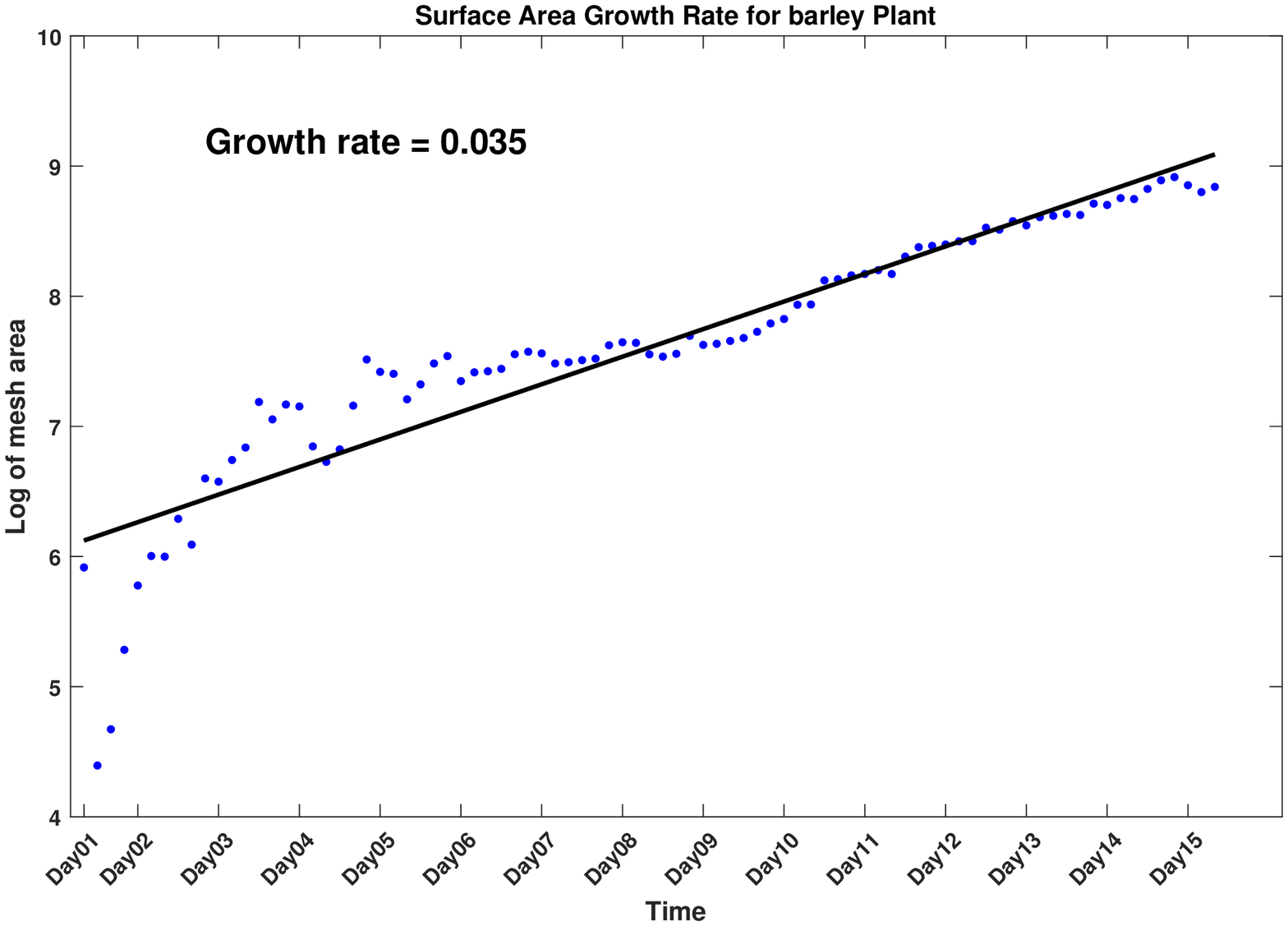} &
\includegraphics[width=3.45in, height=2.25in]{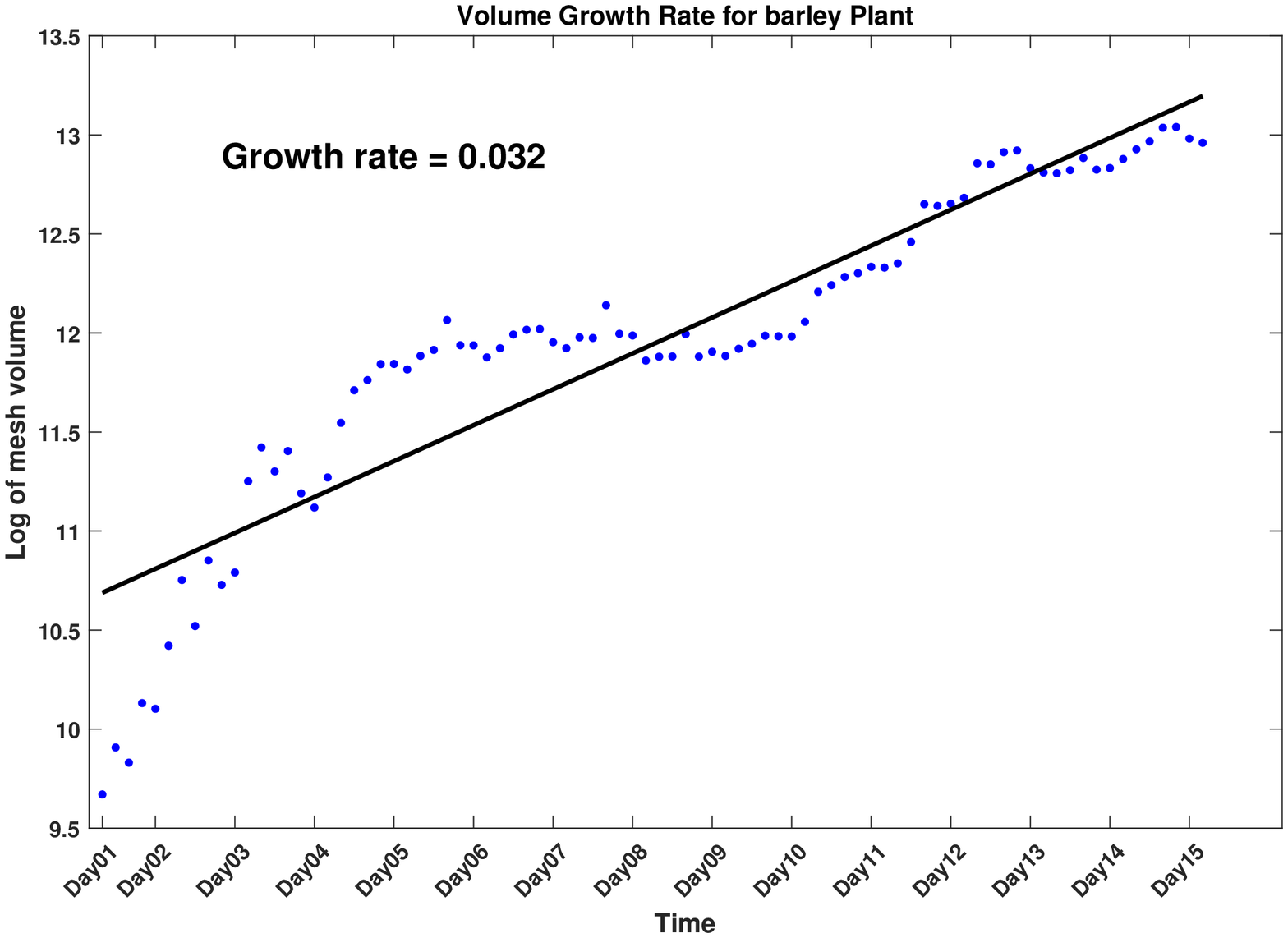} \\
(c) & (d)
\end{tabular}
\end{center}
\caption{(a) and (c): the surface growth rate lines and (c) and (d): the volume growth
rate lines for the Arabidopsis and barley plants.}
\label{growth_rates}
\end{figure*}

The initial short stage of plant growth looks linear, the long intermediate stage of
plant growth looks exponential while the short end stage of plant growth 
looks stationary (or constant). Often, biologists compute the {\bf growth rate} as
the logarithm of mesh surface area values and then fit a straight line to
this data, yielding the maximum exponential growth rate.
Figures \ref{growth_rates}a and \ref{growth_rates}b
show the surface area and volume growth rates for the Arabidopsis plant
while Figures \ref{growth_rates}c and \ref{growth_rates}d
show the surface area and volume growth rates for the barley plant.
The growth rates (slopes of the growth rate lines) are printed
as text items in the upper left corner of each graph and show that
the surface area and volume growth rates for the two plants are roughly the same.
However, as the growth rate of Barley plant exhibits a highly non-linear pattern, 
fitting a straight line to compute the actual growth rate may not be appropriate. 
We believe instead that a polynomial curve fitting scheme 
might be a better, with local growth rates being the slopes
of the tangents on this curve.

\section{Limitations of the System and Future Work}
\label{limitations}

We have presented a fully automated system capable of analyzing plant growth 
throughout the lifetime of the plant. We have validated the
accuracy of the system by experimenting with two real plants throughout their
lifetime, which clearly shows the diurnal nature of plant growth. This type
of system can be used for comparing growth patterns of different types,
varieties and species in different environmental conditions in
autonomous way. The robot can also be used to perform tracking of different plant
organs over time. We believe that the proposed system is general enough to
perform various biological relevant experiments in a non-invasive and automatic 
manner.

\begin{figure}[ht!]
\begin{center}
\includegraphics[width=3.0in]{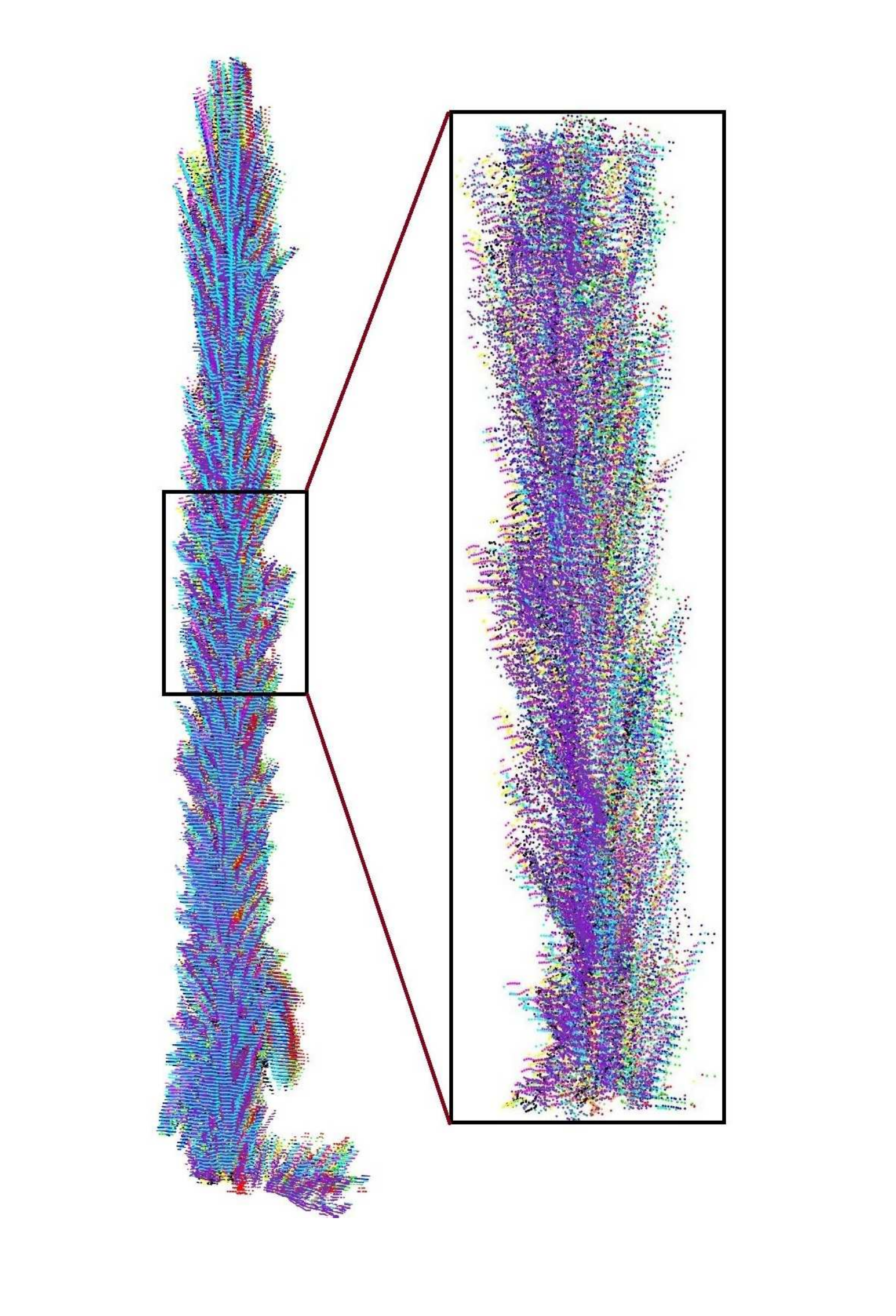}
\caption{Reconstruction of a conifer plant (different colors indicate
different scans).}
\label{conifer}
\end{center}
\end{figure}

\begin{figure}[ht!]
\begin{center}
\includegraphics[width=3.0in]{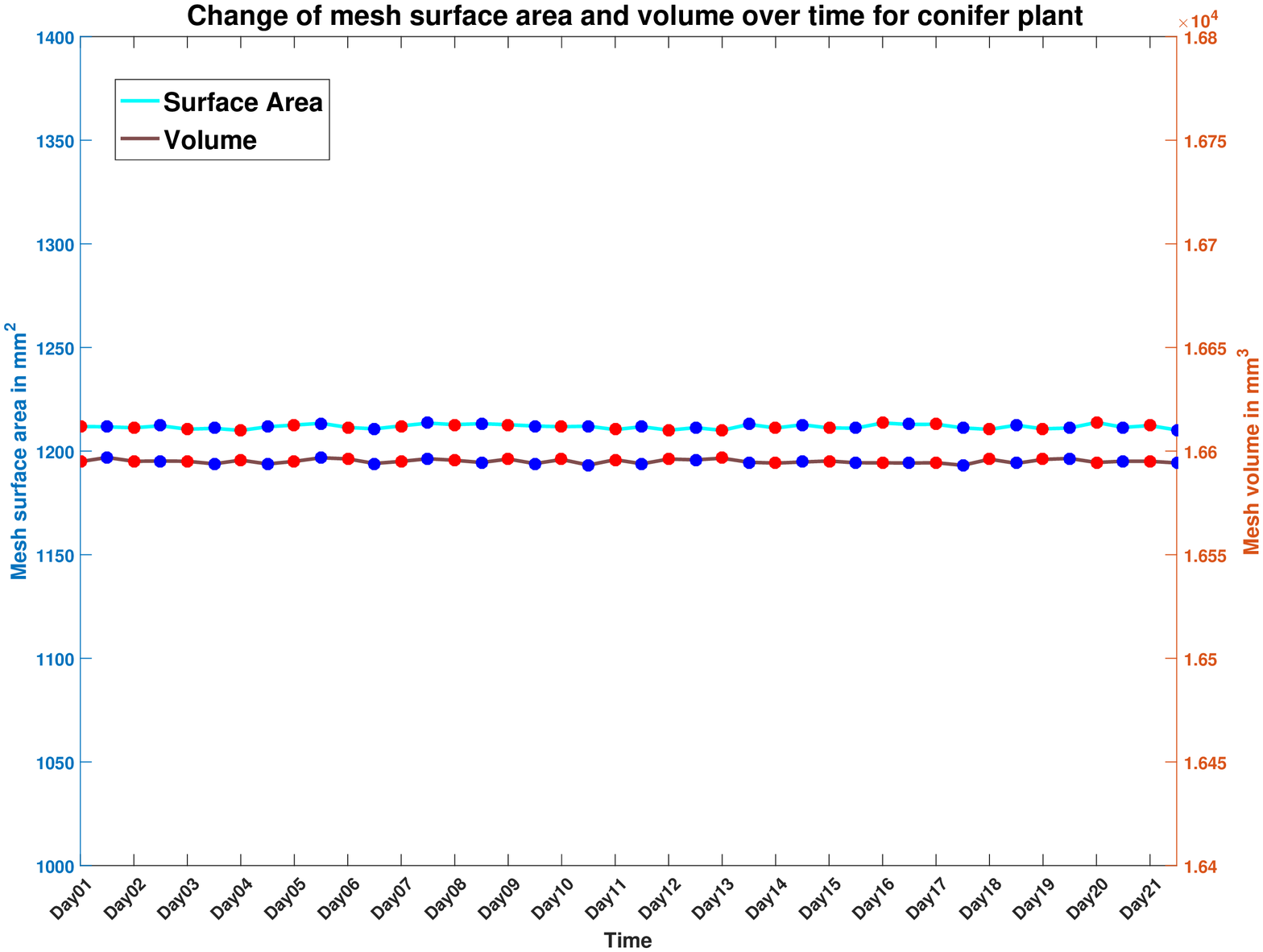}
\caption{Growth data for the conifer plant. The conifer as scanned at 14:00 in the afternoon (light)
and at 2:00 in the night (dark).}
\label{conifer_growth_rate}
\end{center}
\end{figure}

One minor limitation of the system is that, it
can process one plant at a time, but this can easily be changed for
future scanning. To increase plant throughput, we can
scan multiple plants at a time. We have also begun to study more challenging
plants like the conifer. Accurate reconstruction of 
the conifer is challenging due to the fine structures of its needles. 
For example, Figure \ref{conifer} shows the reconstruction of
12 view of a conifer tree with a single cutout shows one part of its surface
at higher resolution. We can see that the range data does not capture
the needle structure adequately. Ideally, each needle of the conifer should be 
clearly and completely visible in the reconstructed point cloud but this is not the case. 
The initial growth pattern of conifer plant is shown in Figure 
\ref{conifer_growth_rate}. The growth rate calculated from this pattern is a flat 
horizontal line (not
shown here) with a slope of 0.0 (effectively, plant growth cannot be captured at the sampling rate of
twice a day we are currently using). Lower sampling rates, for example, once per week might 
capture a growth pattern but certainly not any nightly diurnal growth patterns.
One of the reasons to measure this plant's growth  was to see if we could observe a 
diurnal growth pattern (it is currently unknown if one exists).
Obviously, this is not possible with our current setup.
It is unknown if the needles shrink and expand from night to day (and if they do, can we capture
this information?).  Perhaps, some simple open/closing morphological operations 
would be helpful here.

Leafy plants will also be problematic for our system as the volume measured for
the plant and its actual volume will be very different.
One idea to handle
occlusion can be to exploit the $7$ degrees of freedom of our highly 
flexible robot arm and move it to the occluded areas dynamically to
perform scanning more efficiently. Such areas might be found quickly by
processing the grayvalue images found by the scanning. Such scanning
would have to be performed locally and not on Sharcnet to allow dynamic
re-adjustment of the scanner's trajectory.

\ifCLASSOPTIONcompsoc
  \section*{Acknowledgments}
\else
  \section*{Acknowledgment}
\fi

The authors gratefully acknowledge CFI support to acquire/
build the growth chamber, the robot arm and the near infrared
scanner. The authors also acknowledge financial
support through NSERC discovery grants. H\"uner is grateful
for the financial support of the Canada Research Chair's
programme.

\ifCLASSOPTIONcaptionsoff
  \newpage
\fi

%%%%%%%%%%%%%%%%%%%%%%%%%%%%%%%%%%%%%%%%%%%%%%%%%%%%%%%%%
\bibliographystyle{IEEEtran}
\bibliography{tcbb}

\begin{thebibliography}{10}
\providecommand{\url}[1]{#1}
\csname url@samestyle\endcsname
\providecommand{\newblock}{\relax}
\providecommand{\bibinfo}[2]{#2}
\providecommand{\BIBentrySTDinterwordspacing}{\spaceskip=0pt\relax}
\providecommand{\BIBentryALTinterwordstretchfactor}{4}
\providecommand{\BIBentryALTinterwordspacing}{\spaceskip=\fontdimen2\font plus
\BIBentryALTinterwordstretchfactor\fontdimen3\font minus
  \fontdimen4\font\relax}
\providecommand{\BIBforeignlanguage}[2]{{%
\expandafter\ifx\csname l@#1\endcsname\relax
\typeout{** WARNING: IEEEtran.bst: No hyphenation pattern has been}%
\typeout{** loaded for the language `#1'. Using the pattern for}%
\typeout{** the default language instead.}%
\else
\language=\csname l@#1\endcsname
\fi
#2}}
\providecommand{\BIBdecl}{\relax}
\BIBdecl

\bibitem{leafTracking-TCBB-2015}
B.~Dellen, H.~Scharr, and C.~Torras, ``Growth signatures of rosette plants from
  time-lapse video,'' \emph{{IEEE/ACM} Transactions on Computational Biology
  and Bioinformatics}, vol.~12, no.~6, pp. 1470--1478, 2015.

\bibitem{diurnal-FPB-2012}
T.~Dornbusch, S.~Lorrain, D.~Kuznetsov, A.~Fortier, R.~Liechti, I.~Xenarios,
  and C.~Fankhauser, ``Measuring the diurnal pattern of leaf hyponasty and
  growth in arabidopsis – a novel phenotyping approach using laser
  scanning,'' \emph{Functional Plant Biology}, vol. 39(11), pp. 860--869, 2012.

\bibitem{DNA-nature-2000}
TheArabidopsisGenomeInitiative, ``Analysis of the genome sequence of the
  flowering plant arabidopsis thaliana,'' \emph{Nature}, vol. 408, no. 6814,
  2000.

\bibitem{Brophy-2015}
M.~Brophy, ``Surface reconstruction from noisy and sparse data,'' Ph.D.
  dissertation, Dept. of Computer Science, University of Western Ontario,
  December 2015.

\bibitem{Brophy-et-al-2015}
M.~Brophy, A.~Chaudhury, S.~S. Beauchemin, and J.~L. Barron, ``A method for
  global nonrigid registration of multiple thin structures,'' in
  \emph{Proceedings of the 12th Conference on Computer and Robot Vision (CRV)},
  2015.

\bibitem{Bucksch-et-al-2013}
A.~Bucksch and K.~Khoshelham, ``Localized registration of point clouds of
  botanic trees,'' \emph{IEEE Geoscience and Remote Sensing Letters}, vol.
  10(3), 2013.

\bibitem{treeGRSL-2014}
G.~Zhou, B.~Wang, and J.~Zhou, ``Automatic registration of tree point clouds
  from terrestrial lidar scanning for reconstructing the ground scene of
  vegetated surfaces,'' \emph{IEEE Geoscience and Remote Sensing Letters},
  vol.~11, no.~9, pp. 1654--1658, 2014.

\bibitem{ayan-crv-2015}
A.~Chaudhury, C.~Ward, A.~Talasaz, A.~G. Ivanov, N.~P.~A. H\"uner,
  B.~Grodzinski, R.~V. Patel, and J.~L. Barron, ``Computer vision based
  autonomous robotic system for {3D} plant growth measurement,'' in \emph{Proc.
  of 12th conference on Computer and Robot Vision(CRV)}, 2015.

\bibitem{plantimaging-renaissance-COPB-2013}
E.~P. Spalding and N.~D. Miller, ``Image analysis is driving a renaissance in
  growth measurement,'' \emph{Current Opinion in Plant Biology}, vol. 16(1),
  pp. 100--104, 2013.

\bibitem{htpheno-BMC-2011}
A.~Hartmann, T.~Czauderna, R.~Hoffmann, N.~Stein, and F.~Schreiber,
  ``{HT}pheno: An image analysis pipeline for high-throughput plant
  phenotyping,'' \emph{{BMC} Bioinformatics}, 2011.

\bibitem{phenoSoftware-2014}
C.~Klukas, D.~Chen, and J.~M. Pape, ``Integrated analysis platform: An
  open-source information system for high-throughput plant phenotyping,''
  \emph{Plant Physiology}, vol. 165, no.~2, pp. 506--518, 2014.

\bibitem{scanalyzer}
scanalyzer{-}HTS,
  \url{http://www.lemnatec.com/products/hardware-solutions/scanalyzer-hts/},
  2016.

\bibitem{Subramanian-et-al-MVA-2013}
R.~Subramanian, E.~Spalding, and N.~Ferrier, ``A high throughput robot system
  for machine vision based plant phenotype studies,'' \emph{Machine Vision and
  Applications}, vol. 24(3), pp. 619--636, 2013.

\bibitem{PaulusManual-2014}
S.~Paulus, H.~Schumann, H.~Kuhlmann, and J.~L\'eon, ``High-precision laser
  scanning system for capturing 3d plant architecture and analysing growth of
  cereal plants,'' \emph{Biosystems Engineering}, vol. 121, 2014.

\bibitem{review-pheno-sensors-2014}
L.~Li, Q.~Zhang, and D.~Huang, ``A review of imaging techniques for plant
  phenotyping,'' \emph{Sensors}, vol.~14, no.~11, 2014.

\bibitem{PhenomicsReview-2011}
R.~T. Furbank1 and M.~Tester, ``Phenomics - technologies to relieve the
  phenotyping bottleneck,'' \emph{Trends in Plant Science}, vol.~16, no.~12,
  pp. 635--644, 2011.

\bibitem{FuturePhenotyping-2013}
F.~Fiorani and U.~Schurr, ``Future scenarios for plant phenotyping,''
  \emph{Annual Review of Plant Biology}, vol.~64, 2013.

\bibitem{plantgrowth-importance-AoB-2008}
T.~Fourcaud, X.~Zhang, A.~Stokes, H.~Lambers, and C.~K\"orner, ``Plant growth
  modelling and applications: The increasing importance of plant architecture
  in growth models,'' \emph{Annals of Botany}, vol. 101, pp. 1053--1063, 2008.

\bibitem{Jimenez-et-al-MVA-2000}
A.~R. Jim\'enez, R.~C. Ru\'iz, and J.~L.~P. Rovira, ``A vision system based on
  a laser range-finder applied to robotic fruit harvesting,'' \emph{Machine
  Vision and Applications}, vol. 11(6), pp. 321--329, 2000.

\bibitem{Paulus-et-al-bmc-2013}
S.~Paulus, J.~Dupuis, A.~K. Mahlein, and H.~Kuhlmann, ``Surface feature based
  classification of plant organs from {3D} laserscanned point clouds for plant
  phenotyping,'' \emph{BMC Bioinformatics}, vol.~14, no. 238, 2013.

\bibitem{Paulus-et-al-sensors-2014}
S.~Paulus, J.~Dupuisemail, S.~Riedelemail, and H.~Kuhlmann, ``Automated
  analysis of barley organs using {3D} laser scanning: An approach for high
  throughput phenotyping,'' \emph{Sensors}, vol. 14(7), pp. 12\,670--12\,686,
  2014.

\bibitem{grapevine-BMC-2015}
M.~Klodt, K.~Herzog, R.~T\"opfer, and D.~Cremers, ``Field phenotyping of
  grapevine growth using dense stereo reconstruction,'' \emph{{BMC}
  Bioinformatics}, vol.~16, p. 143, 2015.

\bibitem{organseg-BMC-2015}
M.~Wahabzada, S.~Paulus, K.~Kersting, and A.~Mahlein, ``Automated
  interpretation of 3d laserscanned point clouds for plant organ
  segmentation,'' \emph{{BMC} Bioinformatics}, vol.~16, p. 248, 2015.

\bibitem{Paproki-et-al-bmc-2012}
A.~Paproki, X.~Sirault, S.~Berry, R.~Furbank, and J.~Fripp, ``A novel mesh
  processing based technique for {3D} plant analysis,'' \emph{BMC Plant
  Biology}, vol.~12, no.~1, p.~63, 2012.

\bibitem{Golbach-et-al-MVA-2015}
F.~Golbach, G.~Kootstra, S.~Damjanovic, G.~Otten, and R.~Zedde, ``Validation of
  plant part measurements using a 3{D} reconstruction method suitable for
  high-throughput seedling phenotyping,'' \emph{Machine Vision and
  Applications}, 2015.

\bibitem{segmentationStudy-MVA-2016}
H.~Scharr, M.~Minervini, A.~P. French, C.~Klukas, D.~M. Kramer, X.~Liu,
  I.~Luengo, J.~Pape, G.~Polder, D.~Vukadinovic, X.~Yin, and S.~A. Tsaftaris,
  ``Leaf segmentation in plant phenotyping: a collation study,'' \emph{Machine
  Vision and Applications}, vol.~27, no.~4, pp. 585--606, 2016.

\bibitem{tof-ICRA-2011}
G.~Aleny{\`{a}}, B.~Dellen, and C.~Torras, ``3d modelling of leaves from color
  and tof data for robotized plant measuring,'' in \emph{{IEEE} International
  Conference on Robotics and Automation (ICRA)}, 2011.

\bibitem{Kelly-et-al-MVA-2016}
D.~Kelly, A.~Vatsa, W.~Mayham, and T.~Kazic, ``Extracting complex lesion
  phenotypes in zea mays,'' \emph{Machine Vision and Applications}, vol.~27,
  no.~1, pp. 145--156, 2016.

\bibitem{tomato-PRL-2011}
G.~Xu, F.~Zhang, S.~G. Shah, Y.~Ye, and H.~Mao, ``Use of leaf color images to
  identify nitrogen and potassium deficient tomatoes,'' \emph{Pattern
  Recognition Letters}, vol.~32, no.~11, pp. 1584--1590, 2011.

\bibitem{leaf-CEA-2015}
E.~E. Aksoy, A.~Abramov, F.~W\"org\"otter, H.~Scharr, A.~Fischbach, and
  B.~Dellen, ``Modeling leaf growth of rosette plants using infrared stereo
  image sequences,'' \emph{Computers and Electronics in Agriculture}, vol. 110,
  pp. 78--90, 2015.

\bibitem{Pound-eccv-2014}
M.~P. Pound, A.~P. French, E.~H. Murchie, and T.~P. Pridmore, ``Surface
  reconstruction of plant shoots from multiple views,'' in \emph{Proc. of ECCV
  Workshops}, 2014.

\bibitem{Pound-et-al-MVA-2016}
M.~P. Pound, A.~P. French, J.~A. Fozard, E.~H. Murchie, and T.~P. Pridmore, ``A
  patch-based approach to 3{D} plant shoot phenotyping,'' \emph{Machine Vision
  and Applications}, 2016.

\bibitem{phenotyping-ECCV-2014}
T.~T. Santos, L.~V. Koenigkan, J.~G.~A. Barbedo, and G.~C. Rodrigues, ``3{D}
  plant modeling: Localization, mapping and segmentation for plant phenotyping
  using a single hand-held camera,'' in \emph{Proc. of ECCV Workshops}, 2014.

\bibitem{odometry-MVA-2016}
T.~T. Santos and G.~C. Rodrigues, ``Flexible three-dimensional modeling of
  plants using low-resolution cameras and visual odometry,'' \emph{Machine
  Vision and Applications}, vol.~27, no.~5, pp. 695--707, 2016.

\bibitem{stem-diameter-1-1986}
E.~Garnier and A.~Berger, ``Effect of water stress on stem diameter changes of
  peach trees growing in the field,'' \emph{Journal of Applied Ecology}, vol.
  23(1), pp. 193--209, 1986.

\bibitem{stem-diameter-2-1992}
T.~Simonneau, R.~Habib, J.~Goutouly, and J.~Huguet, ``Diurnal changes in stem
  diameter depend upon variations in water content: Direct evidence in peach
  trees,'' \emph{Journal of Experimental Botany}, vol. 44(3), pp. 615--621,
  1992.

\bibitem{stem-diameter-3-2001}
T.~Genard, S.~Fishman, G.~Vercambre, J.~Huguet, C.~Bussi, J.~Besset, and
  R.~Habib, ``A biophysical analysis of stem and root diameter variations in
  woody plants,'' \emph{Plant Physiology}, vol. 126(2), 2001.

\bibitem{plantimaging-tps-2014}
R.~Sozzani, W.~Busch, E.~P. Spalding, and P.~N. Benfey, ``Advanced imaging
  techniques for the study of plant growth and development,'' \emph{Trends in
  Plant Science}, vol.~19, no.~15, 2014.

\bibitem{circadian-sensors-2012}
P.~J. Navarro, C.~Fern\'andez, J.~Weiss, and .~Marcos, ``Development of a
  configurable growth chamber with a computer vision system to study circadian
  rhythm in plants,'' \emph{Sensors}, vol.~12, no.~11, p. 15356, 2012.

\bibitem{cea-2014}
L.~Benoit, D.~Rousseau, E.~Belin, D.~Demilly, and F.~Chapeau-Blondeau,
  ``Simulation of image acquisition in machine vision dedicated to seedling
  elongation to validate image processing root segmentation algorithms,''
  \emph{Computers and Electronics in Agriculture}, vol. 104, 2014.

\bibitem{Barron-et-al-1994}
J.~Barron and A.~Liptay, ``Optic flow to measure minute increments in plant
  growth,'' \emph{BioImaging}, vol.~2, no.~1, pp. 57--61, 1994.

\bibitem{Liptay-et-al-1995}
A.~Liptay, J.~L. Barron, T.~Jewett, and I.~V. Wesenbeeck, ``Oscillations in
  corn seedling growth as measured by optical flow,'' \emph{Journal of the
  American Society for Horticultural Science}, vol. 120, no.~3, 1995.

\bibitem{treeModel-TCBB-2010}
C.~Godin and P.~Ferraro, ``Quantifying the degree of self-nestedness of trees:
  Application to the structural analysis of plants,'' \emph{{IEEE/ACM}
  Transactions on Computational Biology and Bioinformatics}, vol.~7, no.~4, pp.
  688--703, 2010.

\bibitem{Augustin-et-al-MVA-2015}
M.~Augustin, Y.~Haxhimusa, W.~Busch, and W.~G. Kropatsch, ``A framework for the
  extraction of quantitative traits from 2{D} images of mature arabidopsis
  thaliana,'' \emph{Machine Vision and Applications}, 2015.

\bibitem{4d-plant-siggraph-2013}
Y.~Li, X.~Fan, N.~J. Mitra, D.~Chamovitz, D.~Cohen-Or, and B.~Chen, ``Analyzing
  growing plants from 4{D} point cloud data,'' \emph{ACM Transactions on
  Graphics (Proceedings of SIGGRAPH Asia)}, 2013.

\bibitem{Chui-Rangarajan-2000}
H.~Chui and A.~Rangarajan, ``A feature registration framework using mixture
  models,'' in \emph{Proc. of the IEEE Workshop on Mathematical Methods in
  Biomedical Image Analysis}, ser. MMBIA, 2000.

\bibitem{Tsin-Kanade-2004}
Y.~Tsin and T.~Kanade, ``A correlation-based approach to robust point set
  registration,'' in \emph{Proc. of 8th European Conference on Computer Vision
  (ECCV)}, 2004, pp. 558--569.

\bibitem{Zhang-et-al-2012}
J.~Zhang, Z.~Huan, and W.~Xiong, ``An adaptive gaussian mixture model for
  non-rigid image registration,'' \emph{Journal of Mathematical Imaging and
  Vision}, vol.~44, no.~3, pp. 282--294, Nov. 2012.

\bibitem{Somayajula-et-al-2012}
S.~Somayajula, A.~A. Joshi, and R.~M. Leahy, ``Non-rigid image registration
  using gaussian mixture models,'' in \emph{Proceedings of the 5th
  international conference on Biomedical Image Registration}, ser. WBIR, 2012,
  pp. 286--295.

\bibitem{Besl-McKay-92}
P.~J. Besl and N.~D. McKay, ``A method for registration of 3{D} shapes,''
  \emph{IEEE Transactions on Pattern Analysis and Machine Intelligence},
  vol.~14, no.~2, Feb. 1992.

\bibitem{Turk-Levoy-1994}
G.~Turk and M.~Levoy, ``Zippered polygon meshes from range images,'' in
  \emph{Proc. of SIGGRAPH}, 1994.

\bibitem{Huber-Hebert-2003}
D.~F. Huber and M.~Hebert, ``Fully automatic registration of multiple 3{D} data
  sets,'' \emph{Image and Vision Computing}, vol.~21, no.~7, pp. 637--650,
  2003.

\bibitem{Bouaziz-et-al-2013}
S.~Bouaziz, A.~Tagliasacchi, and M.~Pauly, ``Sparce iterative closest point,''
  \emph{Computer Graphics Forum (Symposium on Geometry Processing)}, vol.~32,
  no.~5, pp. 1--11, 2013.

\bibitem{GMM-2011}
B.~Jian and B.~C. Vemuri, ``Robust point set registration using gaussian
  mixture models,'' \emph{IEEE Transactions on Pattern Analysis and Machine
  Intelligence}, vol.~33, no.~8, Aug. 2011.

\bibitem{CPD-2010}
A.~Myronenko and X.~Song, ``Point set registration: Coherent point drift,''
  \emph{IEEE Transactions on Pattern Analysis and Machine Intelligence},
  vol.~32, no.~12, Dec. 2010.

\bibitem{Toldo2010}
R.~Toldo, A.~Beinat, and F.~Crosilla, ``Global registration of multiple point
  clouds embedding the generalized procrustes analysis into an {ICP}
  framework,'' in \emph{Proc. of 3DPVT}, 2010.

\bibitem{ayan-GRSL}
A.~Chaudhury, M.~Brophy, and J.~L. Barron, ``Junction based correspondence
  estimation of plant point cloud data using subgraph matching,'' \emph{IEEE
  Geoscience and Remote Sensing Letters}, vol.~13, no.~8, pp. 1119--1123, 2016.

\bibitem{code-PAMI-2017}
W.~Y. Lin, F.~Wang, M.~M. Cheng, S.~K. Yeung, P.~H.~S. Torr, M.~N. Do, and
  J.~Lu, ``Code: Coherence based decision boundaries for feature
  correspondence,'' \emph{IEEE Transactions on Pattern Analysis and Machine
  Intelligence}, 2017.

\bibitem{DBSCAN-1996}
M.~Ester, H.~Kriegel, J.~Sander, and X.~Xu, ``A density-based algorithm for
  discovering clusters in large spatial databases with noise,'' in \emph{Proc.
  of International Conference on Knowledge Discovery and Data Mining (KDD)},
  1996.

\bibitem{alpha-shape-1983}
H.~Edelsbrunner, D.~Kirkpatrick, and R.~Seidel, ``On the shape of a set of
  points in the plane,'' \emph{IEEE Transactions on Information Theory},
  vol.~29, no.~4, pp. 551--559, 1983.

\bibitem{meshVol-ICIP-2001}
C.~Zhang and T.~Chen, ``Efficient feature extraction for 2{D}/3{D} objects in
  mesh representation,'' in \emph{Proc. of ICIP}, 2001.

\bibitem{diurnal-Nature-2011}
D.~A. Nusinow, A.~Helfer, E.~E. Hamilton, J.~J. King, T.~Imaizumi, T.~F.
  Schultz, E.~M. Farr\'e, and S.~A. Kay, ``The {ELF}4-{ELF}3-{LUX} complex
  links the circadian clock to diurnal control of hypocotyl growth,''
  \emph{Nature}, vol. 475(7356), pp. 398--402, 2011.

\end{thebibliography}

\end{document}